\documentclass[twocolumn,journal]{IEEEtran}
\usepackage[T1]{fontenc}
\usepackage[latin9]{inputenc}
\usepackage{color}
\usepackage{amsmath}
\usepackage{amsthm}
\usepackage{amssymb}
\usepackage[unicode=true,
 bookmarks=true,bookmarksnumbered=true,bookmarksopen=true,bookmarksopenlevel=1,
 breaklinks=false,pdfborder={0 0 0},pdfborderstyle={},backref=false,colorlinks=false]
 {hyperref}
\hypersetup{pdftitle={Your Title},
 pdfauthor={Your Name},
 pdfpagelayout=OneColumn, pdfnewwindow=true, pdfstartview=XYZ, plainpages=false}

\makeatletter

\providecommand{\tabularnewline}{\\}

\theoremstyle{plain}
\newtheorem{thm}{\protect\theoremname}
\theoremstyle{remark}
\newtheorem{rem}[thm]{\protect\remarkname}

\IEEEoverridecommandlockouts
\usepackage[caption=false,font=footnotesize]{subfig}
\usepackage{cite}

\usepackage{graphicx,import}

\makeatother

\providecommand{\remarkname}{Remark}
\providecommand{\theoremname}{Theorem}

\begin{document}
\global\long\def\quat#1{\boldsymbol{#1}}%

\global\long\def\dq#1{\underline{\boldsymbol{#1}}}%

\global\long\def\hp{\mathbb{H}_{p}}%

\global\long\def\dotmul#1#2{\left\langle #1,#2\right\rangle }%

\global\long\def\partialfrac#1#2{\frac{\partial\left(#1\right)}{\partial#2}}%

\global\long\def\totalderivative#1#2{\frac{d}{d#2}\left(#1\right)}%

\global\long\def\mymatrix#1{\boldsymbol{#1}}%

\global\long\def\vecthree#1{\operatorname{v}_{3}\left(#1\right)}%

\global\long\def\vecfour#1{\operatorname{v}_{4}\left(#1\right)}%

\global\long\def\haminuseight#1{\overset{-}{\mymatrix H}_{8}\left(#1\right)}%

\global\long\def\hapluseight#1{\overset{+}{\mymatrix H}_{8}\left(#1\right)}%

\global\long\def\haminus#1{\overset{-}{\mymatrix H}_{4}\left(#1\right)}%

\global\long\def\haplus#1{\overset{+}{\mymatrix H}_{4}\left(#1\right)}%

\global\long\def\norm#1{\left\Vert #1\right\Vert }%

\global\long\def\abs#1{\left|#1\right|}%

\global\long\def\conj#1{#1^{*}}%

\global\long\def\veceight#1{\operatorname{v}_{8}\left(#1\right)}%

\global\long\def\myvec#1{\boldsymbol{#1}}%

\global\long\def\imi{\hat{\imath}}%

\global\long\def\imj{\hat{\jmath}}%

\global\long\def\imk{\hat{k}}%

\global\long\def\dual{\varepsilon}%

\global\long\def\getp#1{\operatorname{\mathcal{P}}\left(#1\right)}%

\global\long\def\getpdot#1{\operatorname{\dot{\mathcal{P}}}\left(#1\right)}%

\global\long\def\getd#1{\operatorname{\mathcal{D}}\left(#1\right)}%

\global\long\def\getddot#1{\operatorname{\dot{\mathcal{D}}}\left(#1\right)}%

\global\long\def\real#1{\operatorname{\mathrm{Re}}\left(#1\right)}%

\global\long\def\imag#1{\operatorname{\mathrm{Im}}\left(#1\right)}%

\global\long\def\spin{\text{Spin}(3)}%

\global\long\def\spinr{\text{Spin}(3){\ltimes}\mathbb{R}^{3}}%

\global\long\def\distance#1#2#3{d_{#1,\mathrm{#2}}^{#3}}%

\global\long\def\distancejacobian#1#2#3{\boldsymbol{J}_{#1,#2}^{#3}}%

\global\long\def\distancegain#1#2#3{\eta_{#1,#2}^{#3}}%

\global\long\def\distanceerror#1#2#3{\tilde{d}_{#1,#2}^{#3}}%

\global\long\def\dotdistance#1#2#3{\dot{d}_{#1,#2}^{#3}}%

\global\long\def\distanceorigin#1{d_{#1}}%

\global\long\def\dotdistanceorigin#1{\dot{d}_{#1}}%

\global\long\def\squaredistance#1#2#3{D_{#1,#2}^{#3}}%

\global\long\def\dotsquaredistance#1#2#3{\dot{D}_{#1,#2}^{#3}}%

\global\long\def\squaredistanceerror#1#2#3{\tilde{D}_{#1,#2}^{#3}}%

\global\long\def\squaredistanceorigin#1{D_{#1}}%

\global\long\def\dotsquaredistanceorigin#1{\dot{D}_{#1}}%

\global\long\def\crossmatrix#1{\overline{\mymatrix S}\left(#1\right)}%

\global\long\def\constraint#1#2#3{\mathcal{C}_{\mathrm{#1},\mathrm{#2}}^{\mathrm{#3}}}%

\global\long\def\si{\text{R1}}%

\global\long\def\lg{\text{R2}}%

\global\long\def\la{a}%

\global\long\def\vecc{\quat p_{c}^{\lg}}%

\global\long\def\dotvecc{\dot{\quat p}_{c}^{\lg}}%

\global\long\def\javecc{\boldsymbol{J}_{c}^{\lg}}%

\global\long\def\vece{\quat p_{e}^{\lg}}%

\global\long\def\dotvece{\dot{\quat p}_{e}^{\lg}}%

\global\long\def\veca{\quat p_{\la}^{\lg}}%

\global\long\def\dotveca{\dot{\quat p}_{\la}^{\lg}}%

\global\long\def\javeca{\boldsymbol{J}_{\la}^{\lg}}%

\global\long\def\vecasi{\quat p_{\si}^{\la}}%

\global\long\def\dotvecasi{\dot{\quat p}_{\si}^{\la}}%

\global\long\def\javecasi{\boldsymbol{J}_{\si}^{\la}}%

\global\long\def\vecsi{\quat t_{\si}^{\lg}}%

\global\long\def\dotvecsi{\dot{\quat t}_{\si}^{\lg}}%

\global\long\def\javecsi{\boldsymbol{J}_{\si}^{\lg}}%

\global\long\def\fworld{\mathcal{F}_{\text{W}}}%

\global\long\def\flg{\mathcal{F}_{\lg}}%

\global\long\def\centerp{\quat p_{c}}%

\global\long\def\edgep{\quat p_{e}}%

\global\long\def\sip{\quat t_{\si}}%

\global\long\def\lgp{\quat t_{\lg}}%

\global\long\def\dotsip{\dot{\quat t}_{\si}}%

\global\long\def\dotlgp{\dot{\quat t}_{\lg}}%

\global\long\def\dotqsi{\dot{\quat q}_{\si}}%

\global\long\def\dotqlg{\dot{\quat q}_{\lg}}%

\global\long\def\dotq{\dot{\quat q}}%

\global\long\def\qsi{\quat q_{\si}}%

\global\long\def\qlg{\quat q_{\lg}}%

\global\long\def\transjasi{\boldsymbol{J}_{t_{1}}}%

\global\long\def\transjalg{\boldsymbol{J}_{t_{2}}}%

\global\long\def\rotatejasi{\boldsymbol{J}_{r_{1}}}%

\global\long\def\rotatejalg{\boldsymbol{J}_{r_{2}}}%

\global\long\def\cscope{\text{C}_{1}}%

\global\long\def\cillu{\text{C}_{2}}%

\global\long\def\csecond{\text{C3}}%

\global\long\def\rcm{\text{\text{R}}}%

\global\long\def\crcm{\text{C}_{\text{\ensuremath{\rcm}}}}%

\global\long\def\retina{\text{\text{r}}}%

\global\long\def\ceyeball{\text{C}_{\retina}}%

\global\long\def\shaft{\text{s}}%

\global\long\def\cshaft{\text{C}_{\shaft}}%

\global\long\def\trocar{\text{tr}}%

\global\long\def\ctrocar{\text{C}_{\trocar}}%

\global\long\def\tip{\text{\text{t}}}%

\global\long\def\ctip{\text{C}_{\tip}}%

\global\long\def\micro{\text{\text{m}}}%

\global\long\def\cmicro{\text{C}_{\micro}}%

\global\long\def\robot{\text{ro}}%

\global\long\def\crobot{\text{C}_{\robot}}%

\global\long\def\joint{\text{j}}%

\global\long\def\cjoint{\text{C}_{\joint}}%

\global\long\def\safedisrcm{d_{\text{\text{RCM,safe}}}}%

\global\long\def\safedisball{d_{\text{\text{retina},safe}}}%

\global\long\def\safedisshaft{d_{\text{shaft,safe}}}%

\global\long\def\safedistrocar{d_{\text{trocar,safe}}}%

\global\long\def\safedistip{d_{\text{\text{tip},safe}}}%

\global\long\def\disjascope{\mymatrix J_{\cscope}}%

\global\long\def\disjaillu{\mymatrix J_{\cillu}}%

\global\long\def\disjasecond{\mymatrix J_{\text{OP}}}%

\global\long\def\disjascopesafe{\mymatrix J_{\cscope,\text{safe}}}%

\global\long\def\disjaillusafe{\mymatrix J_{\cillu,\text{safe}}}%

\global\long\def\disscope{d_{\cscope}}%

\global\long\def\disillu{d_{\cillu}}%

\global\long\def\dissecond{d_{\text{OP}}}%

\global\long\def\dissecondsquare{D_{t_{\lg},\pi}}%

\global\long\def\dotdisscope{\dot{d}_{\cscope}}%

\global\long\def\dotdisillu{\dot{d}_{\cillu}}%

\global\long\def\dotdissecond{\dot{d}_{\text{OP}}}%

\global\long\def\dotdissecondsquare{\dot{D}_{t_{\lg},\pi}}%

\global\long\def\disscopesafe{d_{\cscope,\text{safe}}}%

\global\long\def\disillusafe{d_{\cillu,\text{safe}}}%

\global\long\def\dotdisscopesafe{\dot{d}_{\cscope,\text{safe}}}%

\global\long\def\dotdisillusafe{\dot{d}_{\cillu,\text{safe}}}%

\global\long\def\wsradius{r_{\text{ms}}}%

\global\long\def\thetascope{\theta_{\cscope}}%

\global\long\def\thetascopesafe{\theta_{\cscope,\text{safe}}}%

\global\long\def\thetaillu{\theta_{\cillu}}%

\global\long\def\thetaillusafe{\theta_{\cillu,\text{safe}}}%

\global\long\def\o{\boldsymbol{0}}%

\global\long\def\sidir{\quat l_{\si}}%

\global\long\def\sir{\quat r_{\si}}%

\global\long\def\lgdir{\quat l_{\lg}}%

\global\long\def\lgr{\quat r_{\lg}}%

\global\long\def\dotsidir{\dot{\quat l}_{\si}}%

\global\long\def\dotlgdir{\dot{\quat l}_{\lg}}%

\global\long\def\dotsir{\dot{\quat r}_{\si}}%

\global\long\def\dotlgr{\dot{\quat r}_{\lg}}%

\global\long\def\gainfirst{\eta_{1\text{st}}}%

\global\long\def\gainsecond{\eta}%

\global\long\def\dampingfirst{\lambda_{1\text{st}}}%

\global\long\def\dampingsecond{\lambda_{2\text{nd}}}%

\global\long\def\first{f_{1\text{st}}}%

\global\long\def\fsecond{f_{2\text{nd}}}%

\global\long\def\opplane{\text{\ensuremath{\underbar{\ensuremath{\quat{\pi}}}}}_{\text{OP}}}%

\global\long\def\unitnormal{\quat u_{\pi}}%

\global\long\def\jaunitnormal{\boldsymbol{J}_{\unitnormal}}%

\global\long\def\dotunitnormal{\dot{\quat u}_{\pi}}%

\global\long\def\normal{\quat n_{\pi}}%

\global\long\def\janormal{\boldsymbol{J}_{\normal}}%

\global\long\def\dotnormal{\dot{\quat n}_{\pi}}%

\global\long\def\pplane{\text{\ensuremath{\underbar{\ensuremath{\quat{\pi}}}}}_{\text{planar}}}%

\global\long\def\overlapthreshold{k_{\text{overlap}}}%

\global\long\def\verticalthreshold{k_{\text{vertical}}}%

\global\long\def\shaftdis{d_{\text{shaft}}}%

\global\long\def\tipdis{d_{\text{tip}}}%

\global\long\def\additionaldis{d_{\text{add}}}%

\global\long\def\retinadis{d_{\text{retina}}}%

\global\long\def\restdis{d_{\text{rest}}}%

\global\long\def\alphadis{d_{\alpha}}%

\title{Autonomous Coordinated Control of the Light Guide for Positioning
in Vitreoretinal Surgery}
\author{Yuki~Koyama, Murilo~M.~Marinho, Mamoru~Mitsuishi, and Kanako~Harada\thanks{This
research was funded in part by the New Energy and Industrial Technology
Development Organization (Japan), in part by the ImPACT Program of
the Council for Science, Technology and Innovation (Cabinet Office,
Government of Japan), and in part by the Mori Manufacturing Research
and Technology Foundation.} \thanks{(\emph{Corresponding author:}
Murilo~M.~Marinho)}\thanks{The authors are with the Department
of Mechanical Engineering, the University of Tokyo, Tokyo, Japan.
\texttt{Emails:\{yuuki-koyama581, murilo, mamoru, kanakoharada\}@g.ecc.u-tokyo.ac.jp}.
}}
\maketitle
\begin{abstract}
Vitreoretinal surgery is challenging even for expert surgeons owing
to the delicate target tissues and the diminutive workspace in the
retina. In addition to improved dexterity and accuracy, robot assistance
allows for (partial) task automation. In this work, we propose a strategy
to automate the motion of the light guide with respect to the surgical
instrument. This automation allows the instrument's shadow to always
be inside the microscopic view, which is an important cue for the
accurate positioning of the instrument in the retina. We show simulations
and experiments demonstrating that the proposed strategy is effective
in a 700-point grid in the retina of a surgical phantom. Furthermore,
we integrated the proposed strategy with image processing and succeeded
in positioning the surgical instrument's tip in the retina, relying
on only the robot's geometric information and microscopic images.
\end{abstract}


\section{Introduction}

\IEEEPARstart{V}{itreoretinal} minimally invasive surgical procedures
are routinely conducted manually by surgeons. The diminutive workspace
in the retina is viewed through an ophthalmic microscope. In one hand,
surgeons hold a surgical instrument, such as 30 $\mathrm{mm}$ long
forceps with 1 $\mathrm{mm}$ diameter. In their other hand, surgeons
hold a small lantern, called a light guide, to illuminate the workspace
in the eyeball. In some cases, a chandelier endoilluminator\footnote{A chandelier illuminator is a light source attached to the wall of
the eye and spreads the light in a wider angle.} or the light provided from the microscope are used.

Among challenging vitreoretinal procedures, two of them stand out:
the peeling of the 2.5 $\mathrm{\mu m}$ thick inner limiting membrane
and the cannulation of retinal blood vessels smaller than 100~$\mathrm{\mu m}$
in diameter. The average amplitude of hand tremors is approximately
100$\mathrm{\,\mu m}$\cite{singhPhysiologicalTremorAmplitude2002}
and considerably higher than the accuracy needed for these tasks.
In this sense, these tasks can be considered difficult to perform
manually, even for experienced surgeons.

In this context, several robotic systems have been developed to enhance
operational accuracy and increase patient safety \cite{channaRoboticVitreoretinalSurgery2017}.
These systems are exclusively made for eye surgery and can be classified
into three main categories: hand-held robotic devices \cite{maclachlanMicronActivelyStabilized2012,kimDesignControlFully2021},
cooperatively-controlled systems \cite{uneriNewSteadyhandEye2010,gijbelsInHumanRobotAssistedRetinal2018},
and tele-operated systems \cite{wilsonIntraocularRoboticInterventional2018,heResearchRealizationMasterSlave2018,edwardsFirstinhumanStudySafety2018a,nasseriIntroductionNewRobot2013}.

\begin{figure}[tbh]
\centering
\def\svgwidth{250pt} 
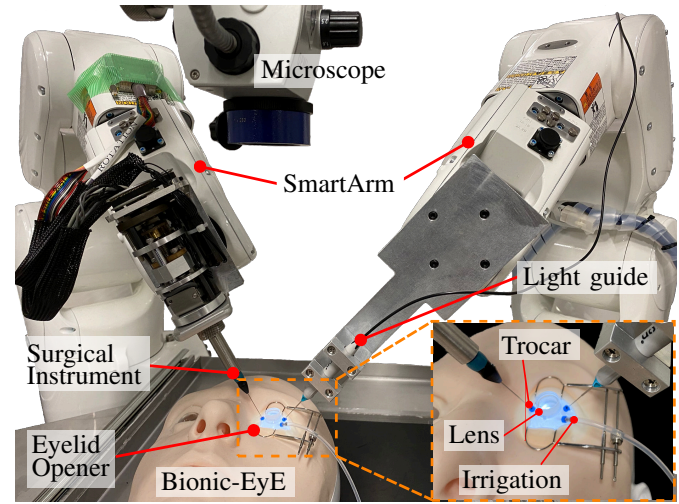

\caption{\label{fig:Surgical_system}The SmartArm \cite{marinhoSmartArmIntegrationValidation2020}
robotic system outfitted for vitreoretinal surgery. The left manipulator
holds a surgical instrument to treat tissues, and the right manipulator
holds a light guide to illuminate the workspace in the eyeball. The
vitreoretinal surgical phantom (Bionic-EyE) \cite{omataSurgicalSimulatorPeeling2018a}
is placed between the two robots.}
\end{figure}

In a different approach to having a robotic system exclusively for
eye surgical procedures, our group has been working on the tele-operated
SmartArm surgical robotic system.\textcolor{blue}{{} }Unlike other tele-operated
systems, the SmartArm system does not have mechanical constraints
for a particular task.\textcolor{blue}{{} }This versatility has allowed
us to preliminarily validate the SmartArm \emph{in-vitro} in multiple
types of surgery in constrained workspaces. For instance, endonasal
surgery \cite{marinhoSmartArmIntegrationValidation2020} and pediatric
surgery \cite{marinhoUnifiedFrameworkTeleoperation2019,marinhoSmartArmSuturingFeasibility2021}.
In another prior work, we have preliminarily established that the
SmartArm can achieve accuracy within the required order of magnitude
for eye surgical procedures, which is about a few dozen micrometers
depending on the task \cite{tomikiUseGeneralpurposeSeriallink2017}.
The average trajectory tracing accuracy of the SmartArm was $22\,\mathrm{\mu m}$
and considerably better than the recorded precision of a surgeon,
which was 108 $\mu m$ \cite{singhPhysiologicalTremorAmplitude2002}.
The SmartArm, equipped for vitreoretinal surgery, is shown in Fig.~\ref{fig:Surgical_system}.

Robotic systems can provide increased dexterity and accuracy and are
designed to enact the motion generated by the surgeon directly. (Semi-)automation
can further improve the safety and efficiency of robot-aided surgical
procedures.

To automate vitreoretinal tasks, accurate and safe positioning of
the surgical instrument's tip on the retina is fundamental. One particular
difficulty of retinal positioning is the estimation of the distance
between the tip of the surgical instrument and the retina. The reduced
size of the workspace, small parallax in stereo microscopes, and the
optical aberrations caused by the vitreous body make regular image
processing algorithms unsuitable. Some groups address this difficulty
by integrating optical coherence tomography (OCT) into their systems
to estimate the instrument's position \cite{ourakCombinedOCTDistance2019a,zhouRoboticEyeSurgery2018}.
However, OCT is high cost, requires an extended processing time, and
has a limited scan range. Some groups have tackled these drawbacks
of OCT, and Sommersperger \emph{et~al.} \cite{sommerspergerRealtimeToolLayer2021}
achieved real-time estimation of the tool to layer distance employing
4D OCT. Nonetheless, those systems are not yet easily available.

In a different approach, our group has been investigating the use
of the instrument's shadow to estimate the proximity between the instrument's
tip and the retina during instrument positioning \cite{tayamaAutonomousPositioningEye2018}.
The rationale behind this idea is that the distance between the instrument's
tip and its shadow and the distance between the instrument's tip and
the retina are proportional related. Therefore, by correctly angling
the lighting instrument, we can use the geometric relationship among
those objects to judge if the positioning has been completed. As a
lighting instrument, we use the light guide given that it allows for
some freedom in controlling the direction of the light.

For this approach to be successful, it is important to ensure that
the tip of the instrument's shadow is visible in the microscopic image
during the positioning process. With this motivation, in this work,
we focus on autonomous bimanual control of the light guide and the
instrument in a further step toward vitreoretinal surgery automation.

\subsection{Related works}

Other groups have also addressed the estimation of the distance or
proximity between the surgical instrument's tip and the retina. For
instance, Richa \emph{et~al}. \cite{richaVisionBasedProximityDetection2012a}
used stereo images and detected the proximity based on the relative
stereo disparity. Bergeless \emph{et~al}. \cite{bergelesSingleCameraFocusBasedLocalization2010,bergelesVisuallyServoingMagnetic2012}
presented a wide-angle localization method for microrobotic devices
using an optical system. The team of Carnegie Mellon University \cite{yangTechniquesRobotaidedIntraocular2018a,routrayRealTimeIncrementalEstimation2019}
proposed a retinal surface estimation method using projected beam
patterns on the retinal surface. In recent research, Kim \emph{et~al}.
\cite{kimAutonomouslyNavigatingSurgical2020} used distinct tool-shadow
dynamics as cues to train their network to automate a tool-navigation
task. Zhou \emph{et~al}. \cite{zhouSpotlightbased3DInstrument2020}
demonstrated a novel method for 3D guidance of the instrument based
on the projection mechanism of a spotlight integrated into the instrument.

Some groups have studied the automation of a single instrument in
eye surgery procedures \cite{beckerSemiautomatedIntraocularLaser2010,xiaMicroscopeGuidedAutonomousClear2020a}.
Other groups have studied bimanual control. Wei \emph{et~al.} \cite{weiweiPerformanceEvaluationMultiarm2009}
evaluated the coupled kinematics of two or more manipulators in hollow
suspended organs, e.g., the eye. More recently, He \emph{et~al.}
\cite{heAutomaticLightPipe2020} proposed an automatic light pipe
actuation system and succeeded in autonomously aligning the light
guide with the target point on the retina.

Existing strategies are not compatible with our shadow-based autonomous
positioning approach \cite{tayamaAutonomousPositioningEye2018}. In
this work, our interest lies in finding a control strategy that guarantees
the visibility of the shadow of the instrument tip during autonomous
positioning. We propose the use of the dynamic regional virtual fixtures
generated by vector-field inequalities (VFIs) \cite{marinhoDynamicActiveConstraints2019}.
Some groups also introduced virtual fixture schemes for robot-aided
vitreoretinal surgery \cite{nasseriVirtualFixtureControl2014,yangSafetyControlMethod2020}.
However, their method needs to artificially reduce the degrees of
freedom to enforce the constraints, and only the constraints around
the insertion point are applied. On the other hand, the VFI method
allows us to satisfy all of the proposed constraints without reducing
the degrees of freedom. This is an important aspect because using
all degrees-of-freedom available for the task incurs, in general,
better dexterity. Moreover, this makes our method practical and independent
of a particular robot's geometry.

The proposed VFIs satisfy \emph{all} the constraints for vitreoretinal
surgery \emph{and} autonomously move the light guide dynamically with
respect to changes in the position of the surgical instrument with
safety guaranteed by design.

\subsection{Statement of contributions\label{subsec:Statement-of-contributions}}

The main contributions of this work are: (1) a novel three-step coordinated
control algorithm for two robotic manipulators aiming shadow-based
autonomous retinal positioning; (2) new vitreoretinal-surgery-specific
dynamic regional virtual fixtures generated by VFIs (Section \ref{sec:Shadow-constraints});
(3) experimental results that demonstrate the real-world feasibility
of our shadow-based autonomous positioning strategy considering, even,
its integration with an image-processing algorithm.

\section{Problem statement\label{sec:Problem-statement}}

\begin{figure}[tbh]
\centering
\def\svgwidth{245pt}
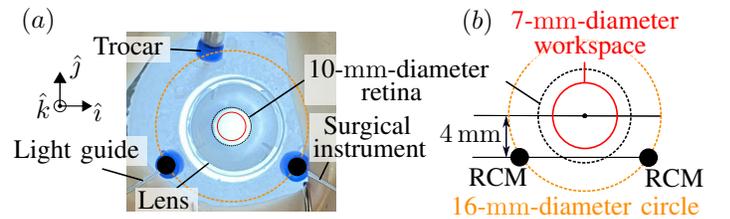

\caption{\label{fig:Workspace_and_RCM}The definition of the workspace and
the location of the RCM. (a) shows the 10-$\mathrm{mm}$-diameter
retina inside the eye phantom which appears smaller because of the
disposable flat lens. The lens has a focal point compatible with the
microscope. (b) shows the definition of the workspace and the RCM
points, which correspond to the trocar points.}
\end{figure}

Consider the setup shown in Fig.~\ref{fig:Surgical_system}. Let
$\si$ be the robot holding the surgical instrument, with joint values
$\qsi$ $\in$ $\mathbb{R}^{n_{1}}$, and $\lg$ be the robot holding
the light guide, with joint values $\qlg$ $\in$ $\mathbb{R}^{n_{2}}$.
The instruments are inserted into a vitreoretinal surgical phantom
(Bionic-EyE \cite{omataSurgicalSimulatorPeeling2018a}) placed between
the two robotic arms. The eye phantom has a 10-$\mathrm{mm}$-diameter
retina, as shown in Fig.~\ref{fig:Workspace_and_RCM}-(a). Similar
to manual surgery, a disposable flat lens is added to provide the
correct view of the workspace. In this work, the size of the workspace
was defined as a 7-$\mathrm{mm}$-diameter circular region in the
retina based on the discussion with partner expert surgeons \cite{sakaiDesignDevelopmentMiniature2014},
as shown in Fig.~\ref{fig:Workspace_and_RCM}-(b)\textcolor{blue}{.}
2D images of the workspace are obtained through an ophthalmic microscope
placed above the eye phantom.

Fig.~\ref{fig:Workspace_and_RCM}-(b) shows the location of the remote
center-of-motion (RCM), which must be generated by the robot to match
the insertion point in the patient. The RCM points correspond to the
trocar points. As in manual surgery, there is some freedom in the
placement of the insertion points. We set the insertion points below
the centerline of the eye for better visibility of the instrument
and its shadow.

\subsection{Goal}

Our goal is to autonomously position the instrument at a point in
the retina using the instrument's shadow as supporting information.
In this work, we propose a reliable control strategy to enact the
envisioned shadow-based retinal positioning approach. Our focus on
this work is the development of the control strategy itself. We integrate
our strategy with image processing for a proof-of-concept validation,
but further progress in that direction is left as a topic for future
work.

\subsection{Premises\label{subsec:Premises}}

Our robot controller was designed under the following premises:
\begin{itemize}
\item The robots are velocity-controlled by the signals $\myvec u_{\text{\text{R1}}}\triangleq\dotqsi\in\mathbb{R}^{n_{1}}$
and $\myvec u_{\text{\text{R2}}}\triangleq\dotqlg\in\mathbb{R}^{n_{2}}$
(Premise~\mbox{I}).
\item The solution $\myvec u_{\text{\text{R1}}}=\mymatrix 0_{n_{1}}$ and
$\myvec u_{\text{\text{R2}}}=\mymatrix 0_{n_{2}}$ must always be
feasible. That is, at worst the robots can stop moving (Premise~\mbox{II}).
\item All constraints (summarized in Section~\ref{subsec:Vitreoretinal-Constraints}
and Section~\ref{subsec:Shadow-Constraints}) must be satisfied at
the beginning of the autonomous motion ($t=0$) (Premise~\mbox{III}).
\item The robots have a kinematic model that is precise enough for the task.
(Premise~\mbox{IV}).
\end{itemize}
Regarding Premise~IV, we have shown in prior work in highly constrained
workspaces that this is a reasonable assumption in terms of safety
\cite{marinhoDynamicActiveConstraints2019}. In terms of precise positioning
in the retina, we have to rely on the microscopic image for increased
accuracy, as we preliminarily do in the experimental section of this
work. In parallel to this work, we are also investigating online-calibration
strategies to adjust the robot's kinematic model online \cite{yoshimuraMBAPoseMaskBoundingBox}.

\subsection{Robot-assisted vitreoretinal task constraints\label{subsec:Vitreoretinal-Constraints}}

A robotic system has to impose the following constraints to safely
conduct vitreoretinal tasks:
\begin{itemize}
\item The shafts of the instruments must always pass through their respective
insertion points (RCM) in the eye ($\crcm$).
\item The light guide's tip must never touch the retina ($\ceyeball$).
\item The shafts of the instruments must never collide with each other ($\cshaft$).
\item The tips of the instruments must always remain inside the eye ($\ctrocar$).
\item The robots must never collide with the microscope ($\cmicro$) and
each other ($\crobot$).
\item The joint values must never exceed their limits ($\cjoint$).
\end{itemize}

\subsection{Shadow-based autonomous positioning constraints\label{subsec:Shadow-Constraints}}

For the shadow-based positioning to guarantee the visibility of the
shadow, we have three additional constraints:
\begin{itemize}
\item The tip of the shadow of the instrument must always be visible in
the microscopic view ($\cscope$).
\item The tip of the instrument must be illuminated sufficiently so that
the contour of its shadow can be visible ($\ctip$) at all times ($\cillu$).
\end{itemize}

\section{Mathematical background}

In this section, we summarize the required mathematical background
for understanding the proposed controller.

\subsection{Quaternions and operators\label{subsec:Quaternions-and-operators}}

The quaternion set is
\[
\mathbb{H}\triangleq\left\{ h_{1}+\imi h_{2}+\imj h_{3}+\imk h_{4}\,:\,h_{1},h_{2},h_{3},h_{4}\in\mathbb{R}\right\} ,
\]
in which the imaginary units $\imi$, $\imj$, and $\imk$ have the
following properties: $\hat{\imath}^{2}=\hat{\jmath}^{2}=\hat{k}^{2}=\hat{\imath}\hat{\jmath}\hat{k}=-1$.
Elements of the set $\mathbb{H}_{p}\triangleq\left\{ \quat h\in\mathbb{H}\,:\,\real{\quat h}=0\right\} $
represent points in $\mathbb{R}^{3}$. The set of quaternions with
unit norm, $\mathbb{S}^{3}\triangleq\left\{ \quat r\in\mathbb{H}\,:\,\norm{\quat r}=1\right\} $,
represent the rotation $\quat r=\cos(\phi/2)+\quat v\sin(\phi/2)$,
where $\phi\in\mathbb{R}$ is the rotation angle around the rotation
axis $\quat v\in\mathbb{S}^{3}\cap\mathbb{H}_{p}$.

The operator $\text{\ensuremath{\mathrm{vec}_{4}}}$ maps a quaternion
$\quat h$ $\in$ $\mathbb{H}$ into a column vector $\mathbb{R}^{4}$.
We use a shorthand version $\text{\ensuremath{\mathrm{v}_{4}}}$ in
the following sections. The Hamilton operators $\overset{+}{\mymatrix H}_{4}$
and $\overset{-}{\mymatrix H}_{4}$ \cite[Def. 2.1.6]{adornoRobotKinematicModeling2017}
satisfy $\vecfour{\quat h\quat h^{\prime}}=\haplus{\quat h}\vecfour{\quat h^{\prime}}=\haminus{\quat h^{\prime}}\vecfour{\quat h}$
for $\quat h,\,\quat h^{\prime}\in\mathbb{H}$. $\mymatrix C_{4}=\text{diag}(1,\,-1,\,-1,\,-1)$
satisfies $\vecfour{\conj{\quat h}}=\mymatrix C_{4}\vecfour{\quat h}$
for $\quat h\in\mathbb{H}$. Furthermore, $\overline{\mymatrix S}$
\cite[Eq. (3)]{marinhoDynamicActiveConstraints2019} has the following
properties $\vecfour{\quat h\times\quat h^{\prime}}=\crossmatrix{\quat h}\vecfour{\quat h^{\prime}}=\crossmatrix{\quat h^{\prime}}^{T}\vecfour{\quat h}$
for $\quat h,\,\quat h^{\prime}\in\mathbb{H}$.

\subsection{Constrained optimization algorithm\label{subsec:Constrained-optimization-algorithm}}

The end effector's tip of two robots can be controlled by a centralized
kinematic control strategy \cite{marinhoUnifiedFrameworkTeleoperation2019}.
In this work, we only need to control the translation of the instruments'
tips because of their symmetry with respect to their shafts. If rotational
control is also required, the objective function in \cite{marinhoUnifiedFrameworkTeleoperation2019}
can be used.

Let $\quat t_{\text{R}i},\quat t_{\text{R}i,d}\in\mathbb{H}_{p}$
be the translation and the desired translation of the end effector
of each $i$-th robot, with $i\in\left\{ 1,2\right\} $. Given $\myvec q=\left[\begin{array}{cc}
\qsi^{T} & \qlg^{T}\end{array}\right]^{T}$, the desired control signal, $\myvec u=\begin{bmatrix}\myvec u_{R1}^{T} & \myvec u_{R2}^{T}\end{bmatrix}^{T}$,
is obtained as
\begin{align}
\myvec u\in\underset{\dotq}{\text{argmin}}\  & \beta\left(f_{t,1}+f_{\lambda,1}\right)+(1-\beta)\left(f_{t,2}+f_{\lambda,2}\right)\label{eq:constrained-optimization-algorithm}\\
\text{subject to} & \ \boldsymbol{W}\dotq\preceq\boldsymbol{w},\nonumber 
\end{align}
in which $f_{t,i}\triangleq\norm{\mymatrix J_{t_{i}}\dot{\myvec q}_{\text{R}i}+\eta\vecfour{\tilde{\quat t}_{i}}}_{2}^{2}$
and $f_{\lambda,i}\triangleq\lambda\norm{\dot{\myvec q}_{\text{R}i}}_{2}^{2}$
are the cost functions related to the end-effector translation and
joint velocities of the $i\text{-th}$ robot. Furthermore, each robot
has a translation Jacobian $\mymatrix J_{t_{i}}\in\mathbb{R}^{4\times n_{i}}$
\cite[ Eq. (6)]{marinhoDynamicActiveConstraints2019} that satisfies
$\vecfour{\dot{\quat t}_{\text{R}i}}=\boldsymbol{J}_{t_{i}}\dot{\quat q}_{\text{R}i}$
and a translation error $\tilde{\quat t}_{i}\triangleq\tilde{\quat t}_{i}\left(\myvec q_{\text{R}i}\right)=\quat t_{\text{R}i}-\quat t_{\text{R}i,d}$.
In addition, $\eta\in(0,\infty)\subset\mathbb{R}$ is a tunable gain,
$\lambda\in[0,\infty)\subset\mathbb{R}$ is a damping factor, and
$\beta\in[0,\,1]\subset\mathbb{R}$ is a weight that defines a ``soft''
priority between the two robots. A ``soft'' priority means that
there is a relative level of importance between robots, defined by
$\beta$, allowing for a good level of customization instead of a
strict hierarchical scheme.

The $r$ inequality constraints $\boldsymbol{W}\dotq\preceq\boldsymbol{w}$,
in which $\boldsymbol{W}\triangleq\boldsymbol{W}\left(\myvec q\right)\in\mathbb{R}^{r\times\left(n_{1}+n_{2}\right)}$,
$\boldsymbol{w}\triangleq\boldsymbol{w}\left(\myvec q\right)\in\mathbb{R}^{r}$
are used to generate active constraints using the VFI method \cite{marinhoDynamicActiveConstraints2019}.
Problem~\eqref{eq:constrained-optimization-algorithm} in general
does not have an analytical solution and a numerical solver must be
used. The Lyapunov stability of such approaches has been proven in
\cite{goncalvesParsimoniousKinematicControl2016}, as long as the
objective function is convex and $\myvec u=0$ is in the feasible
set.

\subsection{Vector-field-inequalities method\label{subsec:VFI-method}}

The vector-field-inequalities (VFI) method \cite{marinhoDynamicActiveConstraints2019}
is used to map task-space constraints into configuration-space constraints.
The VFI method has been shown to be particularly useful for dynamic-active
constraints \cite[Section III]{marinhoVirtualFixtureAssistance2020}.

The VFI method relies on obtaining functions $d\triangleq d(\quat q,t)\in\mathbb{R}$
that represent the signed distance between two geometric primitives.
The time-derivative of the distance is
\begin{align*}
\dot{d}= & \underbrace{\partialfrac{d\left(\quat q,t\right)}{\quat q}}_{\boldsymbol{J}_{d}}\dot{\quat q}+\zeta\left(t\right)\text{,}
\end{align*}
where $\boldsymbol{J}_{d}\in\mathbb{R}^{1\times n}$ is the distance
Jacobian and $\zeta\left(t\right)=\dot{d}-\boldsymbol{J}_{d}\dot{\quat q}$
is the residual, that contains the distance dynamics unrelated to
$\dotq$. Moreover, let there be a safe distance $d_{\text{safe}}\triangleq d_{\text{safe}}\left(t\right)\in[0,\infty)$
and an error $\tilde{d}\triangleq d_{\text{safe}}-d$ to generate
safe zones or $\tilde{d}\triangleq\tilde{d}\left(\quat q,t\right)=d-d_{\text{safe}}$
to generate restricted zones.

With these definitions, and given $\eta_{d}\in[0,\infty)$, the signed
distance dynamics is constrained by $\dot{\tilde{d}}\geq-\eta_{d}\tilde{d}$
in both cases, which actively filters the robot motion only in the
direction approaching the boundary between the primitives so that
the primitives do not collide.

The following constraint is used to generate safe zones, such as the
entry-point constraint, 
\begin{align}
\boldsymbol{J}_{d}\dot{\quat q}\leq & \eta_{d}\tilde{d}-\zeta_{\text{safe}}\left(t\right)\text{,}\label{eq:VFI-safe}
\end{align}
for $\zeta_{\text{safe}}\left(t\right)\triangleq\zeta\left(t\right)-\dot{d}_{\text{safe}}$.
Alternatively, restricted zones, such as constraints to prevent collisions,
are generated by
\begin{align}
-\boldsymbol{J}_{d}\dot{\quat q}\leq & \eta_{d}\tilde{d}+\zeta_{\text{safe}}\left(t\right)\text{.}\label{eq:VFI-restrict}
\end{align}

\section{Overview of the proposed control strategy \label{sec:Bimanual-control-strategy}}

\begin{figure*}[t]
\centering
\def\svgwidth{500pt} 
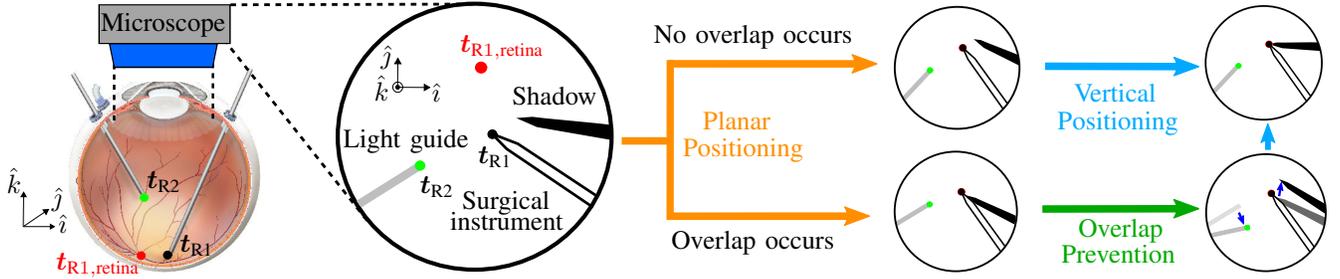

\caption{\label{fig:ConvergingPrevention}The entire process of the proposed
shadow-based autonomous positioning of the surgical instrument's tip
to the retina surface. The proposed method completes positioning in
three steps at most. If the overlap of the surgical instrument's shaft
and its shadow in the microscopic view occurs after the planar positioning
step, the controller removes this overlap in the overlap prevention
step to use the shadow in the vertical positioning step.}
\end{figure*}

Our shadow-based autonomous positioning strategy is divided into three
steps, as described in Fig.~\ref{fig:ConvergingPrevention}: \emph{planar
positioning}, \emph{overlap prevention}, and \emph{vertical positioning}.
Fig.~\ref{fig:Planar_and_vertical_movement} illustrates the planar
and vertical movements of the surgical instrument.

In the planar positioning step described in Section~\ref{sec:Planar-positioning},
the tip of the instrument is moved parallel to the image plane to
a safe point above the target region in the retina. The second step,
overlap prevention, is described in Section\ref{sec:Overlap-prevention}
and ensures that the instrument's tip and its shadow do not overlap
before the third step begins. Finally, in the vertical positioning
step described in Section~\ref{sec:Vertical-positioning}, the instrument
is moved vertically to approach the target point in the retina.

\section{Planar positioning\label{sec:Planar-positioning}}

\begin{figure}[t]
\centering
\def\svgwidth{200pt}
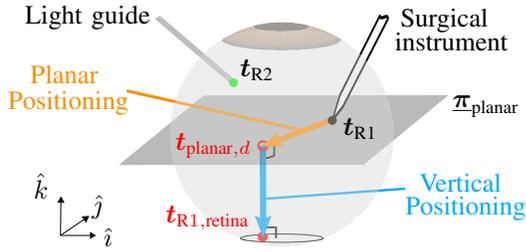

\caption{\label{fig:Planar_and_vertical_movement} The planar and vertical
movements of the tip of the surgical instrument. In the planar positioning
step, the tip is moved parallel to the image plane to a safe point,
$\protect\quat t_{\text{planar},d}$, above the target, $\protect\quat t_{\protect\sip,\text{retina}}$.}
\end{figure}

In the planar positioning step, the controller moves the instrument
from its initial position to $\myvec t_{\text{planar},d}$, which
is the projection of the target position in the retina, $\quat t_{\si,\text{retina}}$,
on the plane $\pplane$, as described in Fig.~\ref{fig:Planar_and_vertical_movement}.
Note that the plane $\pplane$ is parallel to the image plane. To
do so, we use the following control law
\begin{align}
\myvec u\in\underset{\dotq}{\text{argmin}}\  & \beta\left(f_{t,1}+f_{\lambda,1}\right)+(1-\beta)\left(f_{t,2}+f_{\lambda,2}\right)\label{eq:constrained-optimization-algorithm-1}\\
\text{subject to} & \ \begin{bmatrix}\boldsymbol{W}_{\text{vitreo}}\\
\boldsymbol{W}_{\text{shadow}}
\end{bmatrix}\dotq\preceq\begin{bmatrix}\boldsymbol{w}_{\text{vitreo}}\\
\boldsymbol{w}_{\text{shadow}}
\end{bmatrix},\nonumber 
\end{align}
with $\quat t_{\text{R}1,d}=t_{d,\text{plannar}}$, $\quat t_{\text{R}2,d}=\boldsymbol{0}$,
$\beta=0.99$, $\eta=140$, and $\lambda=0.001$. This causes the
light guide to autonomously move in order to keep the constraints
while ``softly'' prioritizing the instrument motion. The inequality
constraints simultaneously enforce the vitreoretinal task constraints
and shadow-based autonomous positioning constraints using VFIs. The
same constraints are used by all steps and are described in detail
in Section~\ref{sec:vitreoretinal-constraints} and Section~\ref{sec:Shadow-constraints}.

This step converges successfully when the error norm goes below a
predefined threshold, i.e., $\norm{\tilde{\quat t}_{1}}\leq0.1\,\mathrm{mm}$,
or fails if the error stops decreasing without error convergence within
the allowed threshold.

\section{Overlap prevention\label{sec:Overlap-prevention}}

After the planar motion finishes successfully, the proposed algorithm
moves on to the overlap prevention step. Preventing the overlap between
the instrument and its shadow can be achieved by moving the tip of
the light guide as far as possible from the plane $\pi_{\text{OP}}$,
which contains the $z-$axis of the world frame and the shaft of the
instrument as described in Fig.~\ref{fig:SecondTaskPlane}, while
keeping the tip of the instrument still. To enact the desired motion,
we use
\begin{align}
\myvec u\in & \underset{\dotq}{\ \text{argmin}}\ \norm{\disjasecond\dotq-\dot{d}_{\text{OP},d}}_{2}^{2}+f_{\lambda,1}+f_{\lambda,2}\label{eq:second_task_optimization}\\
 & \text{\ subject to}\ \begin{array}{l}
\begin{bmatrix}\boldsymbol{W}_{\text{vitreo}}\\
\boldsymbol{W}_{\text{shadow}}
\end{bmatrix}\dotq\preceq\begin{bmatrix}\boldsymbol{w}_{\text{vitreo}}\\
\boldsymbol{w}_{\text{shadow}}
\end{bmatrix}\\
\begin{bmatrix}\mymatrix J_{t_{1}} & \boldsymbol{O}_{4\times n_{2}}\end{bmatrix}\dotq=\boldsymbol{O}_{4\times1}
\end{array},\nonumber 
\end{align}
where $\dissecond(\qsi,\,\qlg)\triangleq\dissecond\in\mathbb{R}$
is the signed distance between the tip of the light guide and the
plane $\pi_{\text{OP}}$ and $\lambda=0.001$. We autonomously control
the light guide and try to increase the signed distance $\dissecond$
at a constant rate $\dot{d}_{\text{OP},d}=0.01\,\mathrm{mm/s}$, which
was determined in pilot experiments. The Jacobian $\disjasecond\in\mathbb{R}^{1\times\left(n_{1}+n_{2}\right)}$
satisfies $\dotdissecond=\disjasecond\dotq$. The distance function,
$\dissecond$, and Jacobian, $\disjasecond$, are described in detail
in Section~\ref{subsec:subtask_signed_distance} and Section~\ref{subsec:subtask_jacobian},
respectively.

Given that it is not possible to know beforehand how far the light
guide can move further from $\pi_{\text{OP}}$, the overlap prevention
step runs until the distance between the shadow's tip and the surgical
instrument's shaft, $\shaftdis$, gets larger than the predefined
threshold, $\overlapthreshold$, in the microscopic view. If the distance
$\shaftdis$ stops increasing but it is not over $\overlapthreshold$,
the controller does not advance to the next step.

\subsection{The signed distance function $\protect\dissecond$\label{subsec:subtask_signed_distance}}

\begin{figure}[t]
\centering
\def\svgwidth{245pt}
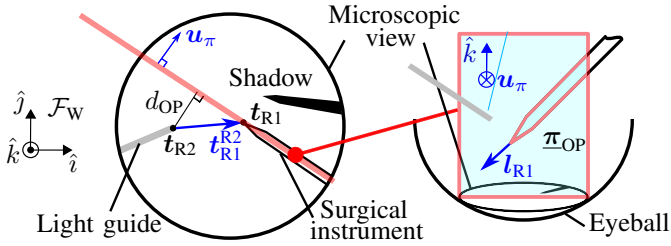

\caption{\label{fig:SecondTaskPlane}The plane, $\protect\opplane$, used to
prevent the overlap of the surgical instrument's shaft and its shadow.
When the light guide's tip is not on $\protect\opplane$, the overlap
does not happen.}
\end{figure}

As shown in Fig.~\ref{fig:SecondTaskPlane}, let the world reference-frame
$\fworld$ be in the retina plane with its $z-$axis pointing upwards,
and define
\begin{align}
\vecsi\left(\qsi,\qlg\right)\triangleq & \vecsi=\sip-\lgp\text{.}\label{eq:t_R2_R1}
\end{align}

Let the rotation of the tip of the surgical instrument be the unit
quaternion $\sir\left(\qsi\right)\triangleq\sir$ $\in$ $\mathbb{S}^{3}$.
In addition, let the unit pure quaternion $\unitnormal\in\mathbb{H}_{p}\cap\mathbb{S}^{3}$
represent the unit normal of the plane. Moreover, let $\sidir\left(\qsi\right)\triangleq\sidir\in\mathbb{H}_{p}\cap\mathbb{S}^{3}$
be the direction of the $z-$axis of the surgical instrument (pointing
outwards, collinear with the shaft of the surgical instrument); that
is, $\sidir=\sir\imk\conj{\sir}$. Then, since the unit plane normal
$\unitnormal$ is orthogonal to both the $z-$axis of the frame $\fworld$,
$\imk$, and the direction of the $z-$axis of the surgical instrument,
$\sidir$, we get\footnote{The unit plane normal $\unitnormal$ is singular when $\sidir\times\imk=0$,
that is, when the surgical instrument's shaft is parallel to the $z-$axis
of the frame $\fworld$. However, this situation never happens inside
the workspace, and this movement is prevented by the constraint $\cmicro$.}
\begin{align}
\normal=\sidir\times\imk,\  & \unitnormal=\frac{\normal}{\norm{\normal}}\text{.}\label{eq:UnitPlaneNormal}
\end{align}

Then, from \eqref{eq:t_R2_R1} and \eqref{eq:UnitPlaneNormal} and
using the inner product, we can get the signed distance $\dissecond$
as follows
\begin{align}
\dissecond= & \dotmul{\unitnormal}{\vecsi}\text{.}\label{eq:SecondDistance}
\end{align}

\subsection{The corresponding Jacobian $\protect\disjasecond$\label{subsec:subtask_jacobian}}

The corresponding Jacobian $\disjasecond$ relates the joint velocities
$\dotq$ to the time derivatives of the signed distance $\dissecond$.
From \eqref{eq:SecondDistance}, the time derivative of the signed
distance is
\begin{align}
\dotdissecond= & \dotmul{\unitnormal}{\dotvecsi}+\dotmul{\vecsi}{\dotunitnormal}\text{.}\label{eq:DistanceSquaredDot}
\end{align}

Then, we have to find $\dotvecsi$ and $\dotunitnormal$ with respect
to $\dotq$. From \eqref{eq:t_R2_R1}, we have
\begin{gather}
\vecfour{\dotvecsi}=\vecfour{\dotsip-\dotlgp}=\underbrace{\begin{bmatrix}\transjasi & -\transjalg\end{bmatrix}}_{\mymatrix J_{\si}^{\lg}}\dotq.\label{eq:Ja_t_R2_R1}
\end{gather}
From \eqref{eq:UnitPlaneNormal} and using the property \cite[Eq. (3)]{marinhoActiveConstraintsUsing2018},
we have
\begin{gather}
\vecfour{\dotunitnormal}=\underbrace{\left(\frac{\boldsymbol{I}_{4\times4}}{\norm{\normal}}-\frac{\vecfour{\normal}\vecfour{\normal}^{T}}{\norm{\normal}^{3}}\right)}_{\quat u_{1}}\vecfour{\dotnormal}\text{.}\label{eq:DotUnit}
\end{gather}

Moreover, 
\begin{gather}
\vecfour{\dotnormal}=\vecfour{\dotsidir\times\imk}\nonumber \\
=\underbrace{\crossmatrix{\imk}^{T}\left(\haminus{\imk\conj{\sir}}+\haplus{\sir\imk}\mymatrix C_{4}\right)\rotatejasi}_{\janormal}\dotqsi\text{,}\label{eq:DotNormal}
\end{gather}
where $\rotatejasi\in\mathbb{R}^{4\times n_{1}}$ is the rotation
Jacobian \cite[ Eq. (7)]{marinhoDynamicActiveConstraints2019}. Substituting
\eqref{eq:DotNormal} into \eqref{eq:DotUnit}, we find
\begin{align}
\vecfour{\dotunitnormal}= & \underbrace{\begin{bmatrix}\quat u_{1}\janormal & \boldsymbol{O}_{4\times6}\end{bmatrix}}_{\jaunitnormal}\dotq\text{.}\label{JaUnitNormal}
\end{align}

Finally, substituting \eqref{eq:Ja_t_R2_R1} and \eqref{JaUnitNormal}
into \eqref{eq:DistanceSquaredDot}, we get
\begin{align}
\dotdissecond= & \underbrace{\left(\vecfour{\unitnormal}^{T}\mymatrix J_{\si}^{\lg}+\vecfour{\vecsi}^{T}\jaunitnormal\right)}_{\disjasecond}\dotq\text{.}\label{eq:JaSecondTask}
\end{align}

\section{Vertical positioning\label{sec:Vertical-positioning}}

After a successful overlap-prevention step, the controller moves on
to the vertical positioning step. The main goal is to get the tip
of the surgical instrument onto the retina precisely, so we prioritize
the motion of the instrument over that of the light guide using a
quadratic programming(QP)-based task-priority framework \cite{kanounKinematicControlRedundant2011}.

To enact the desired motion, we use two cascaded QP optimizations
\begin{align}
\myvec u\in & \underset{\dotq}{\ \text{argmin}}\ \norm{\disjasecond\dotq-\dot{d}_{\text{Desired}}}_{2}^{2}+f_{\lambda,1}+f_{\lambda,2}\label{eq:second_task_optimization-1}\\
 & \text{\ subject to}\ \begin{array}{l}
\begin{bmatrix}\boldsymbol{W}_{\text{vitreo}}\\
\boldsymbol{W}_{\text{shadow}}
\end{bmatrix}\dotq\preceq\begin{bmatrix}\boldsymbol{w}_{\text{vitreo}}\\
\boldsymbol{w}_{\text{shadow}}
\end{bmatrix}\\
\begin{bmatrix}\mymatrix J_{t_{1}} & \boldsymbol{O}_{4\times n_{2}}\end{bmatrix}\dotq=\begin{bmatrix}\mymatrix J_{t_{1}} & \boldsymbol{O}_{4\times n_{2}}\end{bmatrix}\myvec u'
\end{array}\text{,}\nonumber 
\end{align}
where
\begin{align}
\myvec u'\in & \underset{\dotq}{\text{\ argmin}}\ \norm{\mymatrix J_{t_{1}}\dot{\myvec q}_{\si}+\eta\vecfour{\tilde{\quat t}_{1}}}_{2}^{2}+f_{\lambda,1}+f_{\lambda,2}\label{eq:first_task_optimization-1}\\
 & \ \text{subject to}\ \begin{bmatrix}\boldsymbol{W}_{\text{vitreo}}\\
\boldsymbol{W}_{\text{shadow}}
\end{bmatrix}\dotq\preceq\begin{bmatrix}\boldsymbol{w}_{\text{vitreo}}\\
\boldsymbol{w}_{\text{shadow}}
\end{bmatrix},\nonumber 
\end{align}
with $\quat t_{\text{R}1,d}=\quat t_{\text{R}1,\text{retina}}$, $\eta=150$,
and $\lambda=0.0005$. The parameters of \eqref{eq:second_task_optimization-1}
are the same as those of \eqref{eq:second_task_optimization}. This
causes the tip of the instrument to move toward the point in the retina,
while trying, as much as possible, to increase the distance between
the tip of the light guide and $\pi_{\text{OP}}$.

This step converges when the distance between the surgical instrument's
tip and the shadow's tip, $\tipdis$, gets smaller than the predefined
threshold, $\verticalthreshold$, in the microscopic view.

\section{Vitreoretinal task constraints\label{sec:vitreoretinal-constraints}}

\begin{figure}[tbh]
\centering
\def\svgwidth{235pt}
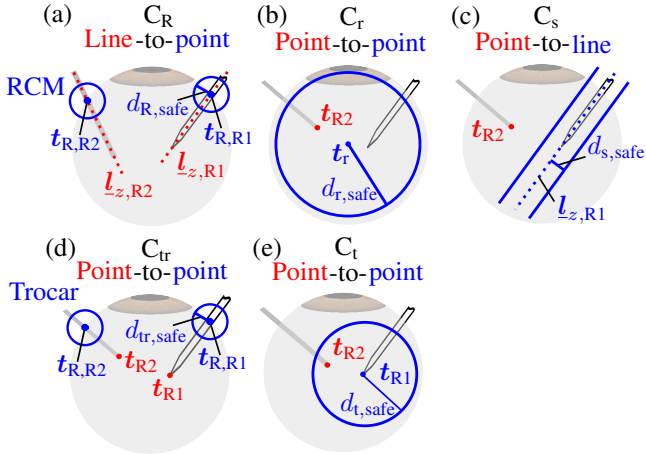

\caption{\label{fig:Vitreoretinal_constraints}The geometrical primitives and
types of VFIs used to generate each task constraint related to the
eye. (a), (b), (c), and (d) show how the vitreoretinal task constraints
$\text{C}_{x}\ (x=\protect\rcm,\,\protect\retina,\,\protect\shaft,\,\protect\trocar)$
are generated. (e) shows how the shadow-based autonomous positioning
constraint $\protect\ctip$ is generated. Note that the trocar points
described in (d) correspond with the RCM points described in (a).}
\end{figure}

\begin{figure}[tbh]
\centering
\def\svgwidth{245pt}
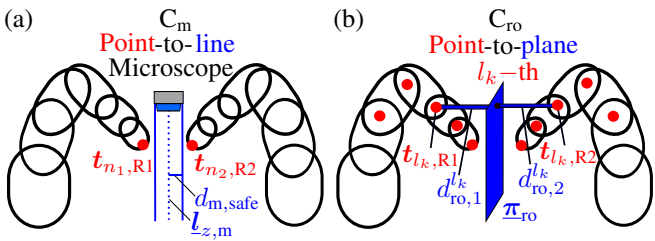

\caption{\label{fig:Robot_constraints}The geometrical primitives and types
of VFIs used to generate each task constraint related to the robots.
(a) and (b) show how the vitreoretinal task constraints $\protect\cmicro$
and $\protect\crobot$ are generated, respectively.}
\end{figure}

This section describes how to enforce the vitreoretinal task constraints
introduced in Section~\ref{subsec:Vitreoretinal-Constraints} using
the VFI method and the Jacobians and distance functions described
in \cite[Section IV]{marinhoDynamicActiveConstraints2019}. Fig.~\ref{fig:Vitreoretinal_constraints}
and Fig.~\ref{fig:Robot_constraints} illustrate the pairs of geometrical
primitives and the type of VFI we use for each pair.

For $\text{C}_{R}$ and $\text{C}_{r}$, based on \eqref{eq:VFI-safe},
we use
\begin{align}
\underbrace{\left[\begin{array}{c}
\begin{array}{cc}
\boldsymbol{J}_{\rcm,1} & \boldsymbol{O}_{1\times n_{2}}\end{array}\\
\begin{array}{cc}
\boldsymbol{O}_{1\times n_{1}} & \boldsymbol{J}_{\rcm,2}\end{array}\\
\begin{array}{cc}
\boldsymbol{O}_{1\times n_{1}} & \boldsymbol{J}_{\retina,2}\end{array}
\end{array}\right]}_{\mymatrix W_{\text{\ensuremath{\rcm},\ensuremath{\retina}}}}\dot{\quat q}\preceq & \underbrace{\left[\begin{array}{c}
\eta_{\rcm}(D_{\rcm\text{,safe}}-D_{\rcm,1})\\
\eta_{\rcm}(D_{\rcm\text{,safe}}-D_{\rcm,2})\\
\eta_{\retina}(D_{\retina\text{,safe}}-D_{\retina,2})
\end{array}\right]}_{\mymatrix w_{\rcm,\retina}}\text{,}\label{eq:Vitreo-VFI-1}
\end{align}
where $D_{\rcm,i}$ and $\boldsymbol{J}_{\rcm,i}$ are the line-to-point
squared distances and Jacobians \cite[Eq.  (33),  (34)]{marinhoDynamicActiveConstraints2019}
between the Plücker lines collinear with the shafts, $\underline{\quat l}_{z,i}$,
and the RCM points, $\quat t_{\rcm,\text{R}i}$. Moreover, $D_{\retina,2}$
and $\boldsymbol{J}_{\retina,2}$ are the point-to-point squared distance
and Jacobian \cite[Eq.  (21),  (22)]{marinhoDynamicActiveConstraints2019}
between the light guide's tip, $\lgp$, and the center point of the
eyeball, $\quat t_{\retina}$. Furthermore, $\eta_{\rcm}=\eta_{\retina}=0.01$,
and $D_{\rcm\text{,safe}}=\left(d_{\rcm\text{,safe}}\right)^{2}=\left(0.5\,\mathrm{mm}\right)^{2}$,
and $D_{\retina\text{,safe}}=\left(d_{\retina\text{,safe}}\right)^{2}=\left(10\,\mathrm{mm}\right)^{2}$.

To enforce $\text{C}_{\text{s}}$\footnote{The possible collision between the light guide and the surgical instrument
is only the collision between the light guide's tip and the surgical
instrument's shaft. This is because the light guide's tip is always
above the surgical instrument's shaft to illuminate the surgical instrument's
tip. This is ensured with the constraint $\cillu$ described in Section
\ref{subsec:Shadow-Constraints}. Therefore, we use a point-to-line
VFI method for the constraint $\cshaft$. Nonetheless, in other scenarios,
a shaft-to-shaft collision avoidance strategy could be used as implemented
in \cite{marinhoDynamicActiveConstraints2019}.}, $\text{C}_{\text{tr}}$, and $\text{C}_{\text{m}}$, based on \eqref{eq:VFI-restrict},
we use
\begin{align}
\underbrace{\left[\begin{array}{c}
\begin{array}{cc}
\boldsymbol{O}_{1\times n_{1}} & -\boldsymbol{J}_{\shaft,2}\end{array}\\
\begin{array}{cc}
-\boldsymbol{J}_{\trocar,1} & \boldsymbol{O}_{1\times n_{2}}\end{array}\\
\begin{array}{cc}
\boldsymbol{O}_{1\times n_{1}} & -\boldsymbol{J}_{\trocar,2}\end{array}\\
\begin{array}{cc}
-\boldsymbol{J}_{\micro,1} & \boldsymbol{O}_{1\times n_{2}}\end{array}\\
\begin{array}{cc}
\boldsymbol{O}_{1\times n_{1}} & -\boldsymbol{J}_{\micro,2}\end{array}
\end{array}\right]}_{\mymatrix W_{\shaft,\trocar,\micro}}\dot{\quat q}\preceq & \underbrace{\left[\begin{array}{c}
\eta_{\shaft}(D_{\shaft,2}-D_{\shaft\text{,safe}})\\
\eta_{\trocar}(D_{\trocar,1}-D_{\trocar\text{,safe}})\\
\eta_{\trocar}(D_{\trocar,2}-D_{\trocar\text{,safe}})\\
\eta_{\micro}(D_{\micro,1}-D_{\micro\text{,safe}})\\
\eta_{\micro}(D_{\micro,2}-D_{\micro\text{,safe}})
\end{array}\right]}_{\mymatrix w_{\shaft,\trocar,\micro}}\text{,}\label{eq:Vitreo-VFI-2}
\end{align}
where $D_{\shaft,2}$ and $\boldsymbol{J}_{\shaft,2}$ are the point-to-line
squared distance and Jacobian \cite[Eq.  (29),  (32)]{marinhoDynamicActiveConstraints2019}
between $\lgp$ and $\underline{l}_{z,1}$, and $D_{\trocar,i}$ and
$\boldsymbol{J}_{\trocar,i}$ are the point-to-point squared distances
and Jacobians between $\quat t_{\text{\text{R}i}}$ and the RCM points,
$\quat t_{\rcm,\text{R}i}$. Moreover, $D_{\micro,i}$ and $\boldsymbol{J}_{\micro,i}$
are the point-to-line squared distances and Jacobians between the
$n_{i}-$th joints of the robots, $\quat t_{n_{i},\text{R}i}$, which
are most likely to collide with the microscope, and the Plücker line
collinear with the microscope direction, $\underline{\quat l}_{z,\micro}$.
Furthermore, $\eta_{\shaft}=\eta_{\micro}=1$, $\eta_{\trocar}=0.01$,
$D_{\shaft\text{,safe}}=\left(d_{\shaft\text{,safe}}\right)^{2}=\left(0.5\,\mathrm{mm}\right)^{2}$,
$D_{\trocar\text{,safe}}=\left(d_{\trocar\text{,safe}}\right)^{2}=\left(5\,\mathrm{mm}\right)^{2}$,
and $D_{\micro\text{,safe}}=\left(6\,\mathrm{cm}\right)^{2}$.

For $\crobot$, we constrain the $l_{k}-$th, $l_{k}\in\left\{ 2,\cdots,n_{i}\right\} $,
joint of the robot $\text{R}_{i}$, $\quat t_{l_{k},\text{R}i}$,
to have a given distance from the plane $\underline{\quat{\text{\ensuremath{\pi}}}}_{\robot}$
described in Fig.~\ref{fig:Robot_constraints}-(b). Let the signed
distances and Jacobians \cite[Eq.  (57),  (59)]{marinhoDynamicActiveConstraints2019}
between $\quat t_{l_{k},\text{R}i}$ and $\underline{\quat{\text{\ensuremath{\pi}}}}_{\robot}$
be $d_{\robot,i}^{l_{k}}\in\mathbb{R}$ and $\boldsymbol{J}_{x,i}^{l_{k}}\in\mathbb{R}^{1\times n_{i}}$,
respectively. We can use
\begin{align}
\underbrace{\left[\begin{array}{c}
\begin{array}{cc}
-\boldsymbol{J}_{\robot,1} & \boldsymbol{O}_{1\times n_{2}}\\
\boldsymbol{O}_{1\times n_{1}} & -\boldsymbol{J}_{\robot,2}
\end{array}\end{array}\right]}_{\mymatrix W_{\text{\ensuremath{\robot}}}}\dot{\quat q}\preceq & \underbrace{\eta_{\robot}\left[\begin{array}{c}
d_{\robot,1}\\
d_{\robot,2}
\end{array}\right]}_{\mymatrix w_{\text{\ensuremath{\robot}}}}\text{,}\label{eq:Vitreo-VFI-3}
\end{align}
where $\eta_{\robot}=1$ and
\begin{align*}
\boldsymbol{J}_{\robot,i}\triangleq\left[\begin{array}{c}
\begin{array}{cc}
\boldsymbol{J}_{\robot,i}^{2} & \boldsymbol{O}_{1\times\left(n_{i}-2\right)}\end{array}\\
\vdots\\
\begin{array}{cc}
\boldsymbol{J}_{\robot,i}^{n_{i}-1} & \boldsymbol{O}_{1\times1}\end{array}\\
\boldsymbol{J}_{\robot,i}^{n_{i}}
\end{array}\right], & d_{\robot,i}\triangleq\left[\begin{array}{c}
d_{\robot,i}^{2}\\
\vdots\\
d_{\robot,i}^{n_{i}-1}\\
d_{\robot,i}^{n_{i}}
\end{array}\right].
\end{align*}

The linear inequality constraints can also be used to enforce joint
limits $\cjoint$ \cite{chengResolvingManipulatorRedundancy1994}
\begin{align}
\underbrace{\left[\begin{array}{c}
\begin{array}{cc}
W_{JL,1} & \boldsymbol{O}_{1\times n_{2}}\end{array}\\
\begin{array}{cc}
\boldsymbol{O}_{1\times n_{1}} & W_{JL,2}\end{array}
\end{array}\right]}_{\mymatrix W_{JL}}\dot{\quat q}\preceq & \underbrace{\left[\begin{array}{c}
w_{JL,1}\\
w_{JL,2}
\end{array}\right]}_{\mymatrix w_{Jl}}\text{,}\label{eq:Vitreo-VFI-4}
\end{align}
where
\begin{align*}
W_{JL,i}\triangleq\left[\begin{array}{c}
-\boldsymbol{I}_{n_{i}\times n_{i}}\\
\boldsymbol{I}_{n_{i}\times n_{i}}
\end{array}\right], & w_{JL,i}\triangleq\left[\begin{array}{c}
\quat q_{\text{min},i}-\quat q_{\text{R}i}\\
\quat q_{\text{max},i}-\quat q_{\text{R}i}
\end{array}\right]\text{,}
\end{align*}
and $\quat q_{\text{min},i},\,\quat q_{\text{max},i}\in\mathbb{R}^{n_{i}}$
are the lower and upper bounds of the joint values.

In conclusion, the vitreoretinal constraints can be enforced using
the following inequality constraint based on \eqref{eq:Vitreo-VFI-1}-\eqref{eq:Vitreo-VFI-4}
\begin{align}
\underbrace{\left[\begin{array}{c}
\mymatrix W_{\text{\ensuremath{\rcm},\ensuremath{\retina}}}\\
\mymatrix W_{\shaft,\trocar,\micro}\\
\mymatrix W_{\text{\ensuremath{\robot}}}\\
\mymatrix W_{\text{limit}}
\end{array}\right]}_{\mymatrix W_{\text{vitreo}}}\dot{\quat q}\preceq & \underbrace{\left[\begin{array}{c}
\mymatrix w_{\text{\ensuremath{\rcm},\ensuremath{\retina}}}\\
\mymatrix w_{\shaft,\trocar,\micro}\\
\mymatrix w_{\text{\ensuremath{\robot}}}\\
\mymatrix w_{\text{limit}}
\end{array}\right]}_{\mymatrix w_{\text{vitreo}}}\text{.}\label{eq:Prior-VFI}
\end{align}

\section{Shadow-based positioning constraints\label{sec:Shadow-constraints}}

This section describes how to enforce the shadow-based autonomous
positioning constraints introduced in Section~\ref{subsec:Shadow-Constraints}.
Firstly, as described in Fig.~\ref{fig:Vitreoretinal_constraints}-(e),
$\ctip$ can be generated using the point-to-point Jacobian $\boldsymbol{J}_{\tip,2}$
and the point-to-point squared distance $D_{\tip,2}$ between $\lgp$
and $\sip$ based on \eqref{eq:VFI-safe}:
\begin{gather}
\underbrace{\begin{bmatrix}\begin{array}{cc}
\boldsymbol{O}_{1\times n_{1}} & \boldsymbol{J}_{\tip,2}\end{array}\end{bmatrix}}_{\mymatrix W_{\tip}}\dotq\preceq\underbrace{\eta_{\tip}(D_{\tip\text{,safe}}-D_{\tip,2})}_{\mymatrix w_{\tip}}\text{,}\label{eq:tip_constraint}
\end{gather}
with $\eta_{\tip}\triangleq0.01$ and $D_{\tip\text{,safe}}=\left(d_{\tip\text{,safe}}\right)^{2}\triangleq\left(10\,\mathrm{mm}\right)^{2}$.

\begin{figure}[tbh]
\centering
\def\svgwidth{245pt}
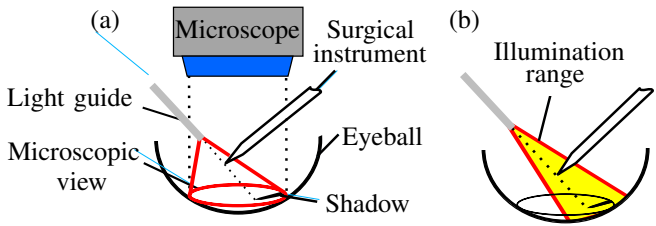

\caption{\label{fig:Suggested_constraints}Conical constraints to enforce $\protect\cscope$
and $\protect\cillu$ needed to guarantee the visibility of the shadow
of the surgical instrument's tip inside the circular microscopic view\@.
(a) In geometrical terms, the tip of the shadow is visible through
the microscopic as long as the surgical instrument's tip is inside
the cone formed by the light guide's tip and the circular microscopic
view. (b) The tip of the surgical instrument is illuminated as long
as the surgical instrument's tip is inside the illumination volume
of the light guide, which is a cone.}
\end{figure}

We can generate $\cscope$ and $\cillu$ by keeping the tip of the
surgical instrument inside the two red cones described in Fig.~\ref{fig:Suggested_constraints}:
the cone formed by the light guide's tip and the circular microscopic
view and the cone which corresponds with the illumination volume of
the light guide.

Quiroz-Omana \emph{et al. }\cite{quiroz-omanaWholeBodyControlSelf2019}
proposed a singularity-free conical constraint based on the VFI for
cones whose central axis never changes. In our use case, the direction
of the central axis shifts depending on the pose of the light guide,
so we propose suitable dynamic conical VFIs that make use of the geometry
of the vitreoretinal task.

Let the signed distances between the surgical instrument's tip and
the time-dependent boundaries of the proposed conical zones be $\tilde{d}_{\cscope}\triangleq\disscope-\disscopesafe\in\mathbb{R}$
and $\tilde{d}_{\cillu}\triangleq\disillu-\disillusafe\in\mathbb{R}$
based on the relationships \eqref{eq:C1_distance_function} and \eqref{eq:C2_distance_function},
respectively. Then, based on \eqref{eq:VFI-restrict}, the constraints
can be implemented as the following linear constraints
\begin{gather}
\underbrace{\begin{bmatrix}-\left(\disjascope-\disjascopesafe\right)\\
-\left(\disjaillu-\disjaillusafe\right)\\
\myvec W_{\tip}
\end{bmatrix}}_{\mymatrix W_{\text{shadow}}}\dotq\preceq\underbrace{\begin{bmatrix}\eta_{1}\left(\disscope-\disscopesafe\right)\\
\eta_{2}\left(\disillu-\disillusafe\right)\\
\myvec w_{\tip}
\end{bmatrix}}_{\mymatrix w_{\text{shadow}}},\label{eq:shadow_constraints}
\end{gather}
where $\eta_{1}=\eta_{2}=0.1$. Furthermore, $\disjascope-\disjascopesafe,\,\disjaillu-\disjaillusafe\in\mathbb{R}^{1\times n}$
are the corresponding Jacobians that relate the joint velocities to
the time derivatives of $\tilde{d}_{\cscope},\,\tilde{d}_{\cillu}$.

\subsection{Distance function and safe distance function for $\protect\cscope$
\label{subsec:Distance-function-C1}}

In this section, the goal is to find a distance function and a safe
distance function\footnote{There are many possible combinations of $\disscope$ and $\disscopesafe$
that can generate $\cscope$. In this work, we discuss one of those
ways.} to enforce $\cscope$. The distance function is related with the
distance between the center of the microscopic view and the tip of
the shadow of the instrument projected on the retina, 
\[
\disscope\left(\qsi,\qlg\right)\triangleq\disscope\in\mathbb{R}^{+}.
\]

The safe distance function delineates the edge of the microscopic
view,
\[
\disscopesafe\left(\qsi,\qlg\right)\triangleq\disscopesafe\in\mathbb{R}^{+}.
\]

\begin{figure}[tbh]
\centering
\def\svgwidth{250pt}
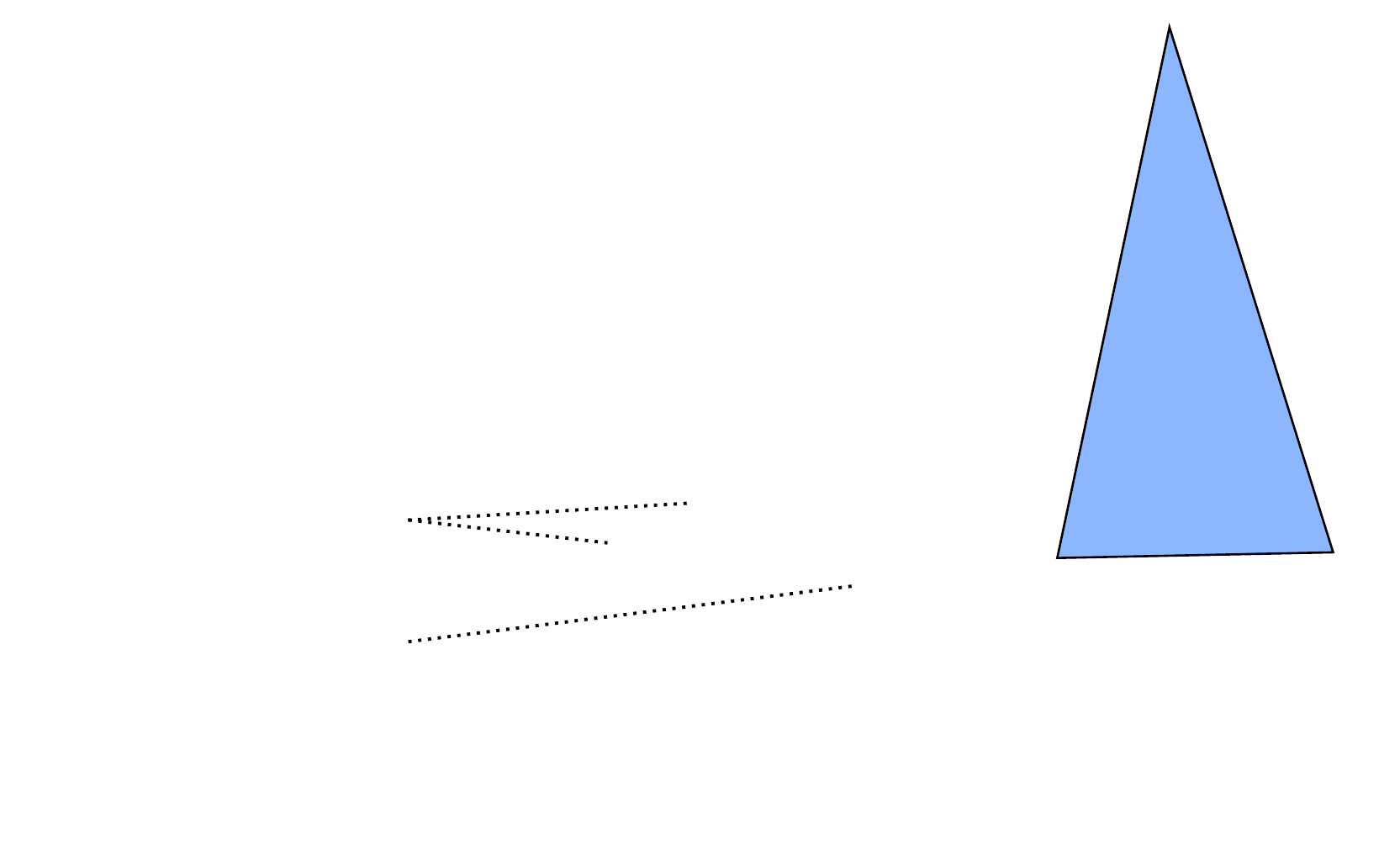

\caption{\label{fig:point_to_cone}The relationship between $\protect\thetascope$
and $\protect\thetascopesafe$ used to define the distance function
$\protect\disscope$ and safe distance function $\protect\disscopesafe$.
When $\protect\thetascope\protect\leq\protect\thetascopesafe$, the
tip of the surgical instrument is kept inside the proposed cone. To
find $\protect\vece$, we have to find $\protect\edgep$, the point
in the edge of the microscopic view and on the plane spanned by $\protect\centerp$,
$\protect\sip$, and $\protect\lgp$.}
\end{figure}

In Fig.~\ref{fig:point_to_cone}, we use the same symbols as in Fig.~\ref{fig:SecondTaskPlane}:
the world reference-frame, $\fworld$, and the translations of the
tips, $\sip$ and $\lgp$. In addition, let $\centerp\in\mathbb{H}_{p}$
represent the center of the circular microscopic view with radius
$\wsradius\in\mathbb{R}^{+}-\left\{ 0\right\} $. Moreover, let $\edgep\left(\qsi,\qlg\right)\triangleq\edgep\in\mathbb{H}_{p}$
be the point in the edge of the microscopic view closest to $\sip$,
hence located on the plane spanned by $\centerp$, $\sip$ and $\lgp$.

To simplify the following explanation, let $\flg$ be the reference
frame with the same orientation as $\fworld$ and whose origin coincides
with $\lgp$. We can re-write $\sip$, $\centerp$ and $\edgep$ with
respect to $\flg$ as \eqref{eq:t_R2_R1}, $\vecc\left(\qlg\right)\triangleq\vecc$
and $\vece\left(\qsi,\qlg\right)\triangleq\vece$, respectively.

With the above definitions, let $\thetascope\left(\qsi,\qlg\right)\triangleq\thetascope\in\left[0,\pi/2\right)\subset\mathbb{R}$
be the angle between $\vecsi$ and $\vecc$; and $\thetascopesafe\left(\qsi,\qlg\right)\triangleq\thetascopesafe\in\left(0,\pi/2\right)$
be the angle between $\vecc$ and $\vece$ as shown in Fig.~\ref{fig:point_to_cone}.
Both the tip of the surgical instrument and its shadow will be visible
in the microscopic view as long as the following inequality holds
\begin{align}
\thetascopesafe & \geq\thetascope.\label{eq:C1_constraint}
\end{align}

To integrate this into the linear constraints \eqref{eq:shadow_constraints},
we need to find Constraint~\eqref{eq:C1_constraint} with respect
to the joint velocities $\quat q$. Since $\vecsi$, $\vecc$ and
$\vece$ are the functions of $\qsi$ and $\qlg$, we propose the
following equivalent constraint
\begin{gather}
\thetascopesafe\geq\thetascope\iff\nonumber \\
\cos^{2}\thetascope-\cos^{2}\thetascopesafe\geq0\iff\nonumber \\
\norm{\vecc}^{2}\norm{\vecsi}^{2}\norm{\vece}^{2}\left(\cos^{2}\thetascope-\cos^{2}\thetascopesafe\right)\geq0.\label{eq:C1_constraint_2}
\end{gather}

We multiplied both sides of \eqref{eq:C1_constraint_2} by $\norm{\vecc}^{2}\norm{\vecsi}^{2}\norm{\vece}^{2}$.
We use squared values to make sure that the time derivatives of the
distance functions, which we need in Subsection \ref{subsec:Distance-Jacobian-C1},
are defined everywhere. Finally, we can remove the explicit dependency
on $\thetascope$ and $\thetascopesafe$ with the dot product between
pure quaternions \cite[Eq. (2)]{marinhoDynamicActiveConstraints2019}
\begin{align}
\dotmul{\vecc}{\vecsi} & =\norm{\vecc}\norm{\vecsi}\cos\thetascope\label{eq:C1_dot_product}\\
\dotmul{\vecc}{\vece} & =\norm{\vecc}\norm{\vece}\cos\thetascopesafe.\label{eq:C1_dot_product_2}
\end{align}

By rearranging Constraint~\eqref{eq:C1_constraint_2} with \eqref{eq:C1_dot_product}
and \eqref{eq:C1_dot_product_2}, we get $\disscope$ and $\disscopesafe$
as follows,
\begin{align}
\norm{\vecc}^{2}\norm{\vecsi}^{2}\norm{\vece}^{2}\left(\cos^{2}\thetascope-\cos^{2}\thetascopesafe\right)\geq0 & \iff\nonumber \\
\underbrace{\norm{\vece}^{2}\left(\dotmul{\vecc}{\vecsi}\right)^{2}}_{\disscope}-\underbrace{\norm{\vecsi}^{2}\left(\dotmul{\vecc}{\vece}\right)^{2}}_{\disscopesafe}\geq0.\label{eq:C1_distance_function}
\end{align}

Note that $\vecsi$ and $\vecc$ can be trivially found as
\begin{align}
\vecsi= & \sip-\lgp & \vecc= & \centerp-\lgp.\label{eq:t_R2_R1_and_p_R2_c}
\end{align}

However, finding $\vece$ requires some geometrical reasoning, as
explained in the following Section.

\subsection{How to find the edge point}

To find $\vece$ as a function of $\qlg$ and $\qsi$, let $\veca\left(\qlg\right)\triangleq\veca$
be the point in $\vecc$ with the same height ($z$-axis coordinate
value) as $\vecsi$. Then, the following relationship holds
\begin{align}
\abs{\dotmul{\veca}{\imk}}= & \abs{\dotmul{\vecsi}{\imk}}\text{.}\label{eq:same_height}
\end{align}

In addition, let $\phi$ $\in$ $\mathbb{R}$ be the angle between
$\vecc$ and the vertical axis, $\imk$, as shown in Fig.~\ref{fig:point_to_cone}.
Since $\phi$ is also the angle between $\veca$ and the vertical
axis, $\imk$, the following holds
\begin{gather}
\cos\phi=\frac{\abs{\dotmul{\veca}{\imk}}}{\norm{\veca}}=\frac{\abs{\dotmul{\vecc}{\imk}}}{\norm{\vecc}}.\label{eq:length_ratio}
\end{gather}

By substituting \eqref{eq:same_height} into \eqref{eq:length_ratio},
we can find 
\begin{gather}
\cos\phi=\frac{\abs{\dotmul{\vecsi}{\imk}}}{\norm{\veca}}=\frac{\abs{\dotmul{\vecc}{\imk}}}{\norm{\vecc}}\implies\frac{\norm{\veca}}{\norm{\vecc}}=\frac{\dotmul{\vecsi}{\imk}}{\dotmul{\vecc}{\imk}},\label{eq:length_ratio_rewritten}
\end{gather}
where we dropped the modulus operator in \eqref{eq:length_ratio_rewritten}
because both inner products will always be negative, given our definition
of $\flg$ and with the assumption that $\lgp$ is higher than both
$\sip$ and $\centerp$ along the vertical axis\footnote{These can be expressed mathematically as $\dotmul{\lgp}{\imk}>\dotmul{\sip}{\imk}\iff\dotmul{\vecsi}{\imk}<0$
and $\dotmul{\lgp}{\imk}>\dotmul{\centerp}{\imk}\iff\dotmul{\vecc}{\imk}<0$,
respectively.}. This is reasonable requirement because, for the shadow of the surgical
instrument to be projected onto the retina, we need that assumption
to hold.

In addition, using the fact that \eqref{eq:length_ratio_rewritten}
represents the ratio of the lengths of $\veca$ and $\vecc$, we can
find
\begin{equation}
\veca=\frac{\norm{\veca}}{\norm{\vecc}}\vecc=\underbrace{\frac{\dotmul{\vecsi}{\imk}}{\dotmul{\vecc}{\imk}}}_{a_{1}}\vecc=a_{1}\vecc.\label{eq:p_R2_a}
\end{equation}

Finally, from the fact that $\vece-\vecc$ and $\veca-\vecsi$ have
the same direction, we find
\begin{gather}
\left(\vece-\vecc\right)=\underbrace{\text{\ensuremath{\frac{\left(\veca-\vecsi\right)}{\norm{\veca-\vecsi}}}}}_{\quat a_{2}}\wsradius\nonumber \\
\iff\vece=\vecc+\quat a_{2}\wsradius.\label{eq:p_R2_e}
\end{gather}

\begin{rem}
\label{rem:singularity1}From \eqref{eq:p_R2_a}, we can see that
there is a singularity when $\dotmul{\vecc}{\imk}=0$, that is, when
the tip of the light guide is on the $x$\textendash $y$ plane of
the retina. However, this is prevented by the vitreoretinal task constraint
$\ceyeball$.
\end{rem}
\begin{rem}
\label{rem:singularity2}As for \eqref{eq:p_R2_e}, there is a singularity
when $\norm{\veca-\vecsi}=0$, which happens when the surgical instrument's
tip, $\sip$, is in the cone's central axis, the line that connects
$\centerp$ and $\lgp$. When we can not define $\disscope-\disscopesafe$
because of this singularity, we, instead, apply another signed distance
calculated using $\norm{\veca-\vecsi}=10\,\mathrm{\mu m}$.
\end{rem}

\subsection{Distance Jacobian and safe distance Jacobian for $\protect\cscope$\label{subsec:Distance-Jacobian-C1}}

Now that we have found $\disscope$and $\disscopesafe$, the next
goal is to find the corresponding Jacobians $\disjascope,\,\disjascopesafe$.
We can get them by finding the time derivatives of $\disscope$ and
$\disscopesafe$ with respect to the joint velocities $\dotq$. The
time derivative of $\disscope$ is
\begin{gather}
\disscope=\overbrace{\norm{\vece}^{2}}^{h_{1}}\overbrace{\left(\dotmul{\vecc}{\vecsi}\right)^{2}}^{h_{2}}\Longrightarrow\dotdisscope=\dot{h}_{1}h_{2}+h_{1}\dot{h}_{2}.\label{eq:dot_C1_distance}
\end{gather}

Analogously, the time derivative of $\disscopesafe$ is
\begin{gather}
\disscopesafe=\overbrace{\norm{\vecsi}^{2}}^{h_{3}}\overbrace{\left(\dotmul{\vecc}{\vece}\right)^{2}}^{h_{4}}\Longrightarrow\dotdisscopesafe=\dot{h}_{3}h_{4}+h_{3}\dot{h}_{4}.\label{eq:dot_C1_safe_distance}
\end{gather}

Moreover, we have (using the property \cite[Eq. (4)]{marinhoDynamicActiveConstraints2019})
\begin{align}
\dot{h}_{1}= & 2\dotmul{\vece}{\dotvece} & \dot{h}_{3}= & 2\dotmul{\vecsi}{\dotvecsi}\label{eq:dot_h1_and_dot_h3}
\end{align}
and
\begin{align}
\dot{h}_{2}= & \overbrace{2\dotmul{\vecc}{\vecsi}}^{h_{5}}\left(\dotmul{\vecsi}{\dotvecc}+\dotmul{\vecc}{\dotvecsi}\right)\label{eq:dot_h2}\\
\dot{h}_{4}= & \underbrace{2\dotmul{\vecc}{\vece}}_{h_{6}}\left(\dotmul{\vece}{\dotvecc}+\dotmul{\vecc}{\dotvece}\right).\label{eq:dot_h4}
\end{align}

Then, we have to find $\dotvecsi$, $\dotvecc$, and $\dotvece$ with
respect to $\dotq$. From \eqref{eq:t_R2_R1_and_p_R2_c}, and by re-writing
with vectors, we have\footnote{$\centerp$ is constant, because the microscope does not move. Therefore,
$\dot{\centerp}=0$} \eqref{eq:Ja_t_R2_R1} and
\begin{gather}
\vecfour{\dotvecc}=\vecfour{-\dotlgp}=\underbrace{\begin{bmatrix}\boldsymbol{O}_{4\times n_{1}} & -\transjalg\end{bmatrix}}_{\mymatrix J_{c}^{\text{\ensuremath{\lg}}}}\dotq\text{.}\label{eq:J_R2_c}
\end{gather}

Lastly, we find $\dotvece$ from \eqref{eq:p_R2_e}
\begin{align}
\dotvece & =\dotvecc+\dot{\quat a}_{2}\wsradius.\label{eq:dot_p_R2_e}
\end{align}

Using the quotient rule, we find
\begin{align}
\dot{\quat a}_{2}=\totalderivative{\frac{\overbrace{\veca-\vecsi}^{\quat h_{7}}}{\underbrace{\norm{\veca-\vecsi}}_{\norm{\quat h_{7}}}}}t= & \frac{\dot{\quat h}_{7}}{\norm{\quat h_{7}}}-\frac{\quat h_{7}\frac{d\left(\norm{\quat h_{7}}\right)}{dt}}{\norm{\quat h_{7}}^{2}}.\label{eq:dot_a2}
\end{align}

Then (using the property \cite[Eq. (3)]{marinhoActiveConstraintsUsing2018}),
we have
\begin{align}
\totalderivative{\norm{\quat h_{7}}}t= & \frac{\dotmul{\quat h_{7}}{\dot{\quat h}_{7}}}{\norm{\quat h_{7}}} & \dot{\quat h}_{7}= & \dotveca-\dotvecsi.\label{eq:dot_h5}
\end{align}

We have found $\dotvecsi$ as a function of $\dotq$ in \eqref{eq:Ja_t_R2_R1}.
Then, the last derivative to be found is that of \eqref{eq:p_R2_a}
\begin{align}
\dotveca= & \dot{a}_{1}\vecc+a_{1}\dotvecc;\label{eq:dot_p_R2_a}
\end{align}
where, using the quotient rule,
\begin{align}
\dot{a}_{1}= & \frac{\dotmul{\imk}{\dotvecsi}}{\dotmul{\vecc}{\imk}}-\frac{\dotmul{\vecsi}{\imk}\dotmul{\imk}{\dotvecc}}{\left(\dotmul{\vecc}{\imk}\right)^{2}}\text{.}\label{eq:dot_a1}
\end{align}

As for \eqref{eq:dot_a1}, using the property \cite[Eq. (2)]{marinhoDynamicActiveConstraints2019}
and from \eqref{eq:Ja_t_R2_R1} and \eqref{eq:J_R2_c}, we get
\begin{align}
\dot{a}_{1}= & \underbrace{\vecfour{\imk}^{T}\left\{ \frac{\mymatrix J_{\text{R1}}^{\text{R2}}}{\dotmul{\vecc}{\imk}}-\frac{\dotmul{\vecsi}{\imk}\mymatrix J_{c}^{\text{R2}}}{\left(\dotmul{\vecc}{\imk}\right)^{2}}\right\} }_{\mymatrix J_{a_{1}}}\dotq.\label{eq:J_a1}
\end{align}

We can now work backwards to find intermediate Jacobians that compose
the distance Jacobians. Substituting \eqref{eq:J_a1} and \eqref{eq:J_R2_c}
in \eqref{eq:dot_p_R2_a} and noticing that $a_{1}\in\mathbb{R},\ \vecc\in\mathbb{H}_{p}$,
we find
\begin{gather}
\vecfour{\dotveca}=\vecfour{\vecc}\dot{a}_{1}+a_{1}\vecfour{\dotvecc}\nonumber \\
=\underbrace{\left\{ \vecfour{\vecc}\mymatrix J_{a_{1}}+a_{1}\mymatrix J_{c}^{\text{R2}}\right\} }_{\mymatrix J_{a}^{\text{R2}}}\dotq\label{eq:J_R2_a}
\end{gather}

Substituting \eqref{eq:J_R2_a} and \eqref{eq:Ja_t_R2_R1} in \eqref{eq:dot_h5}
results in
\begin{align}
\vecfour{\dot{\quat h}_{7}} & =\overbrace{\left(\mymatrix J_{a}^{\text{R2}}-\mymatrix J_{\text{R1}}^{\text{R2}}\right)}^{\mymatrix J_{h_{7}}}\dotq\label{eq:J_h5}\\
\totalderivative{\norm{\quat h_{7}}}t & =\frac{\vecfour{\quat h_{7}}^{T}\vecfour{\dot{\quat h}_{7}}}{\norm{\quat h_{7}}}=\overbrace{\frac{\vecfour{\quat h_{7}}^{T}\mymatrix J_{h_{7}}}{\norm{\quat h_{7}}}}^{\mymatrix J_{\norm{h_{7}}}}\dotq.\label{eq:J_h5_norm}
\end{align}

Moreover, substituting \eqref{eq:J_h5} and \eqref{eq:J_h5_norm}
in \eqref{eq:dot_a2} results in
\begin{gather}
\vecfour{\dot{\quat a}_{2}}=\underbrace{\left(\frac{\mymatrix J_{h_{7}}}{\norm{\quat h_{7}}}-\frac{\vecfour{\quat h_{7}}\mymatrix J_{\norm{h_{7}}}}{\norm{\quat h_{7}}^{2}}\right)}_{\mymatrix J_{a_{2}}}\dotq.\label{eq:J_a2}
\end{gather}

Finally, substituting \eqref{eq:J_a2} and \eqref{eq:J_R2_c} in \eqref{eq:dot_p_R2_e},
we get the last derivative to be found as
\begin{align}
\vecfour{\dotvece} & =\overbrace{\left(\mymatrix J_{c}^{\text{\ensuremath{\lg}}}+r\mymatrix J_{a_{2}}\right)}^{\mymatrix J_{e}^{\lg}}\dotq.\label{eq:J_R2_e}
\end{align}

Then, using $\mymatrix J_{\si}^{\lg},\,\mymatrix J_{c}^{\lg},\,\mymatrix J_{e}^{\lg}$,
we get \eqref{eq:dot_h1_and_dot_h3}, \eqref{eq:dot_h2} and \eqref{eq:dot_h4}
as
\begin{align}
\dot{h}_{1}= & \overbrace{2\vecfour{\vece}^{T}\mymatrix J_{e}^{\text{R2}}}^{\mymatrix J_{h_{1}}}\dotq & \dot{h}_{3}= & \overbrace{2\vecfour{\vecsi}^{T}\mymatrix J_{\text{R1}}^{\text{R2}}}^{\mymatrix J_{h_{3}}}\dotq,\label{eq:J_h1_and_J_h3}
\end{align}
\begin{gather}
\dot{h}_{2}=\overbrace{h_{5}\left\{ \vecfour{\vecsi}^{T}\mymatrix J_{c}^{\text{R2}}+\vecfour{\vecc}^{T}\mymatrix J_{\text{R1}}^{\text{R2}}\right\} }^{\mymatrix J_{h_{2}}}\dotq,\label{eq:J_h2}\\
\dot{h}_{4}=\underbrace{h_{6}\left\{ \vecfour{\vece}^{T}\mymatrix J_{c}^{\text{R2}}+\vecfour{\vecc}^{T}\mymatrix J_{e}^{\text{R2}}\right\} }_{\mymatrix J_{h_{4}}}\dotq.\label{eq:J_h4}
\end{gather}

In conclusion, we can find the distance Jacobian for $\cscope$ by
substituting \eqref{eq:J_h1_and_J_h3} and \eqref{eq:J_h2} in \eqref{eq:dot_C1_distance},
as follows
\begin{gather}
\dot{d}_{C1}=\overbrace{\left(h_{2}\mymatrix J_{h_{1}}+h_{1}\mymatrix J_{h_{2}}\right)}^{\disjascope}\dot{\myvec q}.\label{eq:J_d_C1}
\end{gather}

Besides, we can find the safe distance Jacobian for $\cscope$ by
substituting \eqref{eq:J_h1_and_J_h3} and \eqref{eq:J_h4} in \eqref{eq:dot_C1_safe_distance},
as follows
\begin{gather}
\dotdisscopesafe=\overbrace{\left(h_{4}\mymatrix J_{h_{3}}+h_{3}\mymatrix J_{h_{4}}\right)}^{\disjascopesafe}\dotq.\label{eq:J_d_C1_safe}
\end{gather}

\subsection{Distance function and safe distance function for $\protect\cillu$\label{subsec:Distance-function-C2}}

The goal of this section is to find a distance function and a safe
distance function to enforce $\cillu$. We define these distances
similarly to what was done in Section~\ref{subsec:Distance-function-C1}.
The distance function is related with the distance between the axis
of the shaft of the light guide and the tip of the surgical instrument,
\[
\disillu\left(\qsi,\qlg\right)\triangleq\disillu\in\mathbb{R}^{+}.
\]

The safe distance function corresponds to the edge of the illumination
volume of the light guide,
\[
\disillusafe\left(\qsi,\qlg\right)\triangleq\disillusafe\in\mathbb{R}^{+}.
\]

\begin{figure}[tbh]
\centering
\def\svgwidth{200pt}
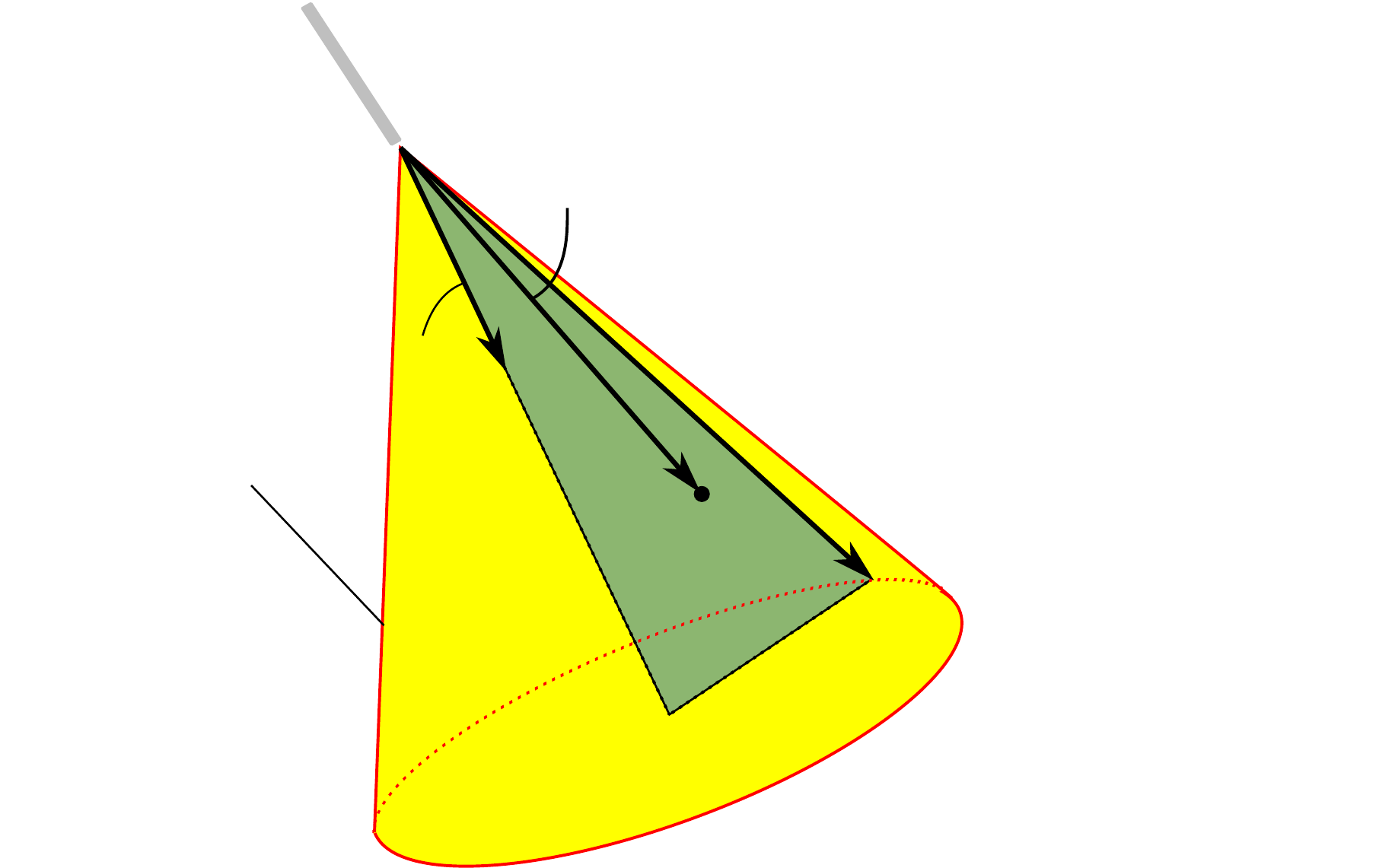

\caption{\label{fig:point_to_illumination}The relationship between $\protect\thetaillu$
and $\protect\thetaillusafe$ used to define the distance function
$\protect\disillu$ and safe distance function $\protect\disillusafe$.
When $\protect\thetaillu\protect\leq\protect\thetaillusafe$, the
tip of the surgical instrument is kept inside the proposed cone.}
\end{figure}

In the following explanation, let the rotation of the light guide
with respect to $\fworld$ be the unit quaternion $\lgr\left(\qlg\right)\triangleq\lgr$
$\in$ $\mathbb{S}^{3}$. In addition, let the unit pure quaternion
$\lgdir\left(\qlg\right)\triangleq\lgdir$$\in$ $\mathbb{H}_{p}\cap\mathbb{S}^{3}$
be the direction of the $z-$axis of the light guide (pointing outwards,
collinear with the shaft of the light guide, the central axis of the
illumination cone); that is,
\begin{equation}
\lgdir=\lgr\imk\conj{\lgr}.\label{eq:l_r2}
\end{equation}

With the above definitions, let $\thetaillu\left(\qsi,\qlg\right)\triangleq\thetaillu$
$\in$ $\left[0,\pi/2\right)$ $\subset$ $\mathbb{R}$ be the angle
between $\lgdir$ and $\vecsi$; and $\thetaillusafe=0.5\,\mathrm{rad}$,
which is constant in time because the illumination volume is inherent
to the light guide, be the angle between $\lgdir$ and the boundary
of the illumination volume. The tip of the surgical instrument will
always be in the illumination volume as long as the following constraint
holds\footnote{Since $\thetaillusafe$ is constant and does not depend on $\qsi$
nor $\qlg$, we do not need to remove the explicit dependency on $\thetaillusafe$.
Therefore, we multiply the following inequality with $\norm{\lgdir}^{2}\norm{\vecsi}^{2}$
to get only the dot product \eqref{eq:C2_dot_product}.}
\begin{gather}
\thetaillusafe\geq\thetaillu\iff\label{eq:C2_constraint}\\
\cos^{2}\thetaillu-\cos^{2}\thetaillusafe\geq0\iff\\
\norm{\lgdir}^{2}\norm{\vecsi}^{2}\left(\cos^{2}\thetaillu-\cos^{2}\thetaillusafe\right)\geq0.\label{eq:C2_constraint_2}
\end{gather}

From the definition of the dot product, we have
\begin{align}
\dotmul{\lgdir}{\vecsi} & =\norm{\lgdir}\norm{\vecsi}\cos\thetaillu.\label{eq:C2_dot_product}
\end{align}

By rearranging Constraint~\eqref{eq:C2_constraint_2} with \eqref{eq:C2_dot_product},
we can remove the explicit dependency on $\thetaillu$, as follows
\begin{gather}
\norm{\lgdir}^{2}\norm{\vecsi}^{2}\left(\cos^{2}\thetaillu-\cos^{2}\thetaillusafe\right)\geq0\iff\nonumber \\
\underbrace{\left(\dotmul{\lgdir}{\vecsi}\right)^{2}}_{\disillu}-\underbrace{\norm{\lgdir}^{2}\norm{\vecsi}^{2}\cos^{2}\thetaillusafe}_{\disillusafe}\geq0.\label{eq:C2_distance_function}
\end{gather}

\subsection{Distance Jacobian and safe distance Jacobian for $\protect\cillu$}

We have found $\disillu$ and $\disillusafe$. We next find their
corresponding Jacobians $\disjaillu,\,\disjaillusafe$. Similarly
to what was done in Section~\ref{subsec:Distance-Jacobian-C1}, we
can get them by finding the time derivatives of $\disillu$ and $\disillusafe$
with respect to the joint velocities $\dotq$. The time derivative
of $\disillu$ is
\begin{align}
 & \disillu=\left(\dotmul{\lgdir}{\vecsi}\right)^{2}\nonumber \\
\Longrightarrow & \dotdisillu=2\dotmul{\lgdir}{\vecsi}\left(\dotmul{\vecsi}{\dotlgdir}+\dotmul{\lgdir}{\dotvecsi}\right)\text{.}\label{eq:dot_C2_distance}
\end{align}

Besides, the time derivative of $\disillusafe$ is
\begin{align}
 & \disillusafe=\overbrace{\norm{\lgdir}^{2}}^{h_{8}}\overbrace{\norm{\vecsi}^{2}}^{h_{9}}\cos^{2}\thetaillusafe\nonumber \\
\Longrightarrow & \dotdisillusafe=\left(\dot{h}_{8}h_{9}+h_{8}\dot{h}_{9}\right)\cos^{2}\thetaillusafe\text{.}\label{eq:dot_c2_safe_distance}
\end{align}

As for $\dot{\quat h}_{8},\,\dot{\quat h}_{9}$, using the property
\cite[Eq. (4)]{marinhoDynamicActiveConstraints2019}, we have
\begin{align}
\dot{h}_{8}= & 2\dotmul{\lgdir}{\dotlgdir} & \dot{h}_{9}= & 2\dotmul{\vecsi}{\dotvecsi}.\label{eq:dot_h6_and_dot_h7}
\end{align}

Then, we need to find $\dotlgdir$ and $\dotvecsi$ with respect to
$\dotq$. We have found $\dotvecsi$ in \eqref{eq:Ja_t_R2_R1}. From
\eqref{eq:l_r2}, as we calculate $\vecfour{\dotsidir}$ in \eqref{eq:DotNormal},
we obtain
\begin{align}
\vecfour{\dotlgdir}= & \overbrace{\left(\haminus{\imk\conj{\lgr}}+\haplus{\lgr\imk}\mymatrix C_{4}\right)}^{\boldsymbol{H}}\rotatejalg\dotqlg\nonumber \\
= & \underbrace{\boldsymbol{H}\left[\begin{array}{cc}
\boldsymbol{O}_{4\times n_{1}} & \rotatejalg\end{array}\right]}_{\boldsymbol{J}_{\lgdir}}\dotq,\label{eq:J_l_R2}
\end{align}
where $\rotatejalg\in\mathbb{R}^{4\times n_{2}}$ is the rotation
Jacobian of the robot $\lg$.

We can now work backwards to find intermediate Jacobians that compose
the distance Jacobians. Substituting \eqref{eq:J_l_R2} and \eqref{eq:Ja_t_R2_R1}
in \eqref{eq:dot_h6_and_dot_h7} and the property \cite[Eq. (2)]{marinhoDynamicActiveConstraints2019}
result in
\begin{align}
\text{\ensuremath{\dot{h}_{8}}}= & \overbrace{2\vecfour{\lgdir}^{T}\boldsymbol{J}_{\lgdir}}^{\boldsymbol{J}_{h_{8}}}\dot{\quat q}\label{eq:dot_h_6}\\
\dot{h}_{9}= & \underbrace{2\vecfour{\vecsi}^{T}\boldsymbol{J}_{\si}^{\lg}}_{\boldsymbol{J}_{h_{9}}}\dot{\quat q}.\label{eq:dot_h_7}
\end{align}

In conclusion, substituting \eqref{eq:Ja_t_R2_R1} and \eqref{eq:J_l_R2}
in \eqref{eq:dot_C2_distance} gives us the distance Jacobian for
$\cillu$, as follows
\begin{gather}
\dotdisillu=\overbrace{2\dotmul{\lgdir}{\vecsi}\left\{ \vecfour{\vecsi}^{T}\boldsymbol{J}_{\lgdir}+\vecfour{\lgdir}^{T}\boldsymbol{J}_{\si}^{\lg}\right\} }^{\disjaillu}\dotq.
\end{gather}

Lastly, we can obtain the safe distance Jacobian for $\cillu$ by
substituting \eqref{eq:dot_h_6} and \eqref{eq:dot_h_7} in \eqref{eq:dot_c2_safe_distance},
as follows
\begin{equation}
\dotdisillusafe=\overbrace{\left(h_{8}\boldsymbol{J}_{h_{9}}+h_{8}\boldsymbol{J}_{h_{9}}\right)\cos^{2}\thetaillusafe}^{\disjaillusafe}\dotq.
\end{equation}

\section{Simulation and Experiment\label{sec:Simulation-and-Experiment}}

We conducted three experiments and performed one simulation study
to evaluate our proposed shadow-based autonomous positioning strategy.

First, in experiment \textbf{E1}, we evaluated the effectiveness and
safety of our proposed conical constraints using a physical robotic
system. Second, in simulation \textbf{S1}, we evaluated the autonomous
instrument positioning in the entire retinal region using our three-step
strategy. Third, in experiment \textbf{E2}, we evaluated a few representative
points obtained in simulation \textbf{S1}. Lastly, in Experiment \textbf{E3,}
we integrated our proposed autonomous control strategy with a data-driven
image-processing algorithm and validated the fully integrated system.

\subsection{System configuration}

The simulations and experiments used the same software implementation
on a Ubuntu 20.04 x64 system. The b-Cap protocol was used to communicate
with the two robot arms (DENSO VS050, DENSO WAVE Inc., Japan) with
a sampling frequency of $150$~Hz. Ros Noetic Ninjemys was used for
the interprocess communication and CoppeliaSim (Coppelia Robotics,
Switzerland) for the simulations. The dual quaternion algebra and
robot kinematics were implemented using DQ Robotics \cite{adornoDQRoboticsLibrary2020}
for Python3.

\subsection{Setup}

The physical robotic setup shown in Fig.~\ref{fig:Surgical_system}
was used for the experiments. A simulated environment replicating
this physical setup was implemented in CoppeliaSim.

A 7-$\mathrm{mm}$-diameter retinal workspace (see Fig.~\ref{fig:Workspace_and_RCM})
inside the Bionic-Eye was used. While performing the tasks described
in the evaluation, the robot was commanded to maintain the shadow
inside the workspace using the proposed conical constraints with $\wsradius=3.5\,\mathrm{mm}$.

Our focus in this work is to propose a safe and reliable robot control
strategy for autonomous positioning on the retina. The image-processing
aspect has been extensively addressed in related works \cite{richaVisualTrackingSurgical2011a,richaVisionBasedProximityDetection2012a},
and in this work, for simplicity, we use the Bionic-Eye with a white
retinal background.

The instruments were initially positioned in such a way that all of
the proposed constraints were satisfied (Premise \mbox{III}).

The calibration between elements in the workspace was performed once
before the beginning of the trials. The calibration of each instrument's
tip with respect to each robot was done through a pivoting process
as often performed in the industry \cite{marinhoDynamicActiveConstraints2019}.
The calibration between the bases of each robot was conducted in a
similar manner, using an image-based position tracker (Polaris Vega,
NDI, Canada). The robots went through a high-accuracy calibration\footnote{This procedure is performed before the robots are shipped. The end-user
has access to those parameters, but the details of the calibration
procedure are patented \cite{denso2017calibration}. Nonetheless,
those parameters do not deviate much from their nominal values.}. Using that information, the calibration between robots and the surgical
phantom was performed using the kinematic models of the robots (Premise
IV) and the positions of the insertion points described in Fig.~\ref{fig:Workspace_and_RCM}.
As routinely performed in robotic surgery, points 5~$\text{mm}$
far from the tip are measured and marked. The instruments are then
inserted into the Bionic-EyE until the mark matches the trocar points
in the Bionic-Eye, by operating the robots' teaching pendants. Lastly,
using the calculated positions of the insertion points and the relationship
described in Fig.~\ref{fig:Workspace_and_RCM}, we found the pose
of the Bionic-EyE in the coordinate of the robot.

\subsection{Distance information in microscopic view\label{subsec:Distance-information}}

Our strategy uses the distance information in the microscopic view
to switch between steps in our three-step algorithm. Namely, we use
the distance between the shadow's tip and the surgical instrument's
shaft, $\shaftdis$, for the overlap prevention step; and the distance
between the shadow's tip and the surgical instrument's tip, $\tipdis$,
for the vertical positioning step.

In the first two experiments with the physical system, these distances
were calculated geometrically, using only the kinematic information
of the robots (Premise \mbox{IV}), and the thresholds $\overlapthreshold$
and $\verticalthreshold$ were set to $0.5\,\mathrm{mm}$ and $0.3\,\mathrm{mm}$,
respectively\footnote{We set the thresholds so that the tip of the surgical instrument stopped
before the tip touched and damaged the retina since image processing
was not integrated in these experiments.}. In the final experiment, we used an image processing strategy to
obtain these distances, and the thresholds $\overlapthreshold$ and
$\verticalthreshold$ were set to $20\,\mathrm{px}$ and $1\,\mathrm{px}$,
respectively. The values of the thresholds were determined for the
movement of the instruments to appear natural in the microscopic image.
Notice that we integrate our strategy with image processing as a feasibility
study of the fully integrated system. There is no claim of novelty
in the imaging-processing part, given that our autonomous control
strategy can be integrated with any suitable imaging-processing strategy.

\subsection{Experiment E1: Evaluation of proposed conical constraints}

The first experiment was conducted to evaluate the conical VFIs we
proposed for the visibility of the shadow.

In this experiment, the surgical instrument's tip was commanded to
follow an arbitrarily predefined trajectory. The light guide was autonomously
controlled by the proposed strategy. The instrument was moved in the
plane, $\pplane$, described in Fig.~\eqref{fig:Planar_and_vertical_movement},
and the constrained optimization problem \eqref{eq:constrained-optimization-algorithm-1}
was used. We defined the trajectory of the surgical instrument's tip
as described in Fig.~\ref{fig:shadow_trajectory_result} so that
the shadow gets outside the 7-$\mathrm{mm}$-diameter workspace when
no constraints are enforced.

We conducted the experiment under two conditions: with and without
the proposed conical constraints. Then, we evaluated the behavior
of the shadow in the microscopic view and the values of the conical
constraints.

\subsubsection*{Results and discussion}

\begin{figure}[tbh]
\centering
\def\svgwidth{240pt}
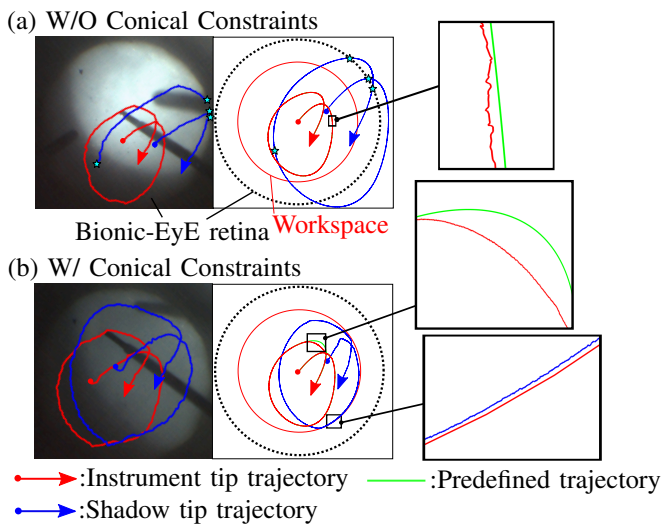

\caption{\label{fig:shadow_trajectory_result}The trajectory results of the
evaluation of the proposed conical VFIs for the shadow visibility.
(a) and (b) show the representative trajectories without and with
the conical constraints, respectively. The controller succeeded in
keeping the tip of the shadow inside the workspace.}
\end{figure}

The resulting trajectories are shown in Fig.~\ref{fig:shadow_trajectory_result}.
The trajectories depicted on the microscopic views were obtained by
manually annotating the experimental videos. The trajectories calculated
using the kinematic models are described next to the microscopic views.
The controller succeeded in maintaining the tip of the shadow inside
the workspace during the entire trial using the proposed conical constraints.

\begin{figure}[tbh]
\centering
\def\svgwidth{250pt}
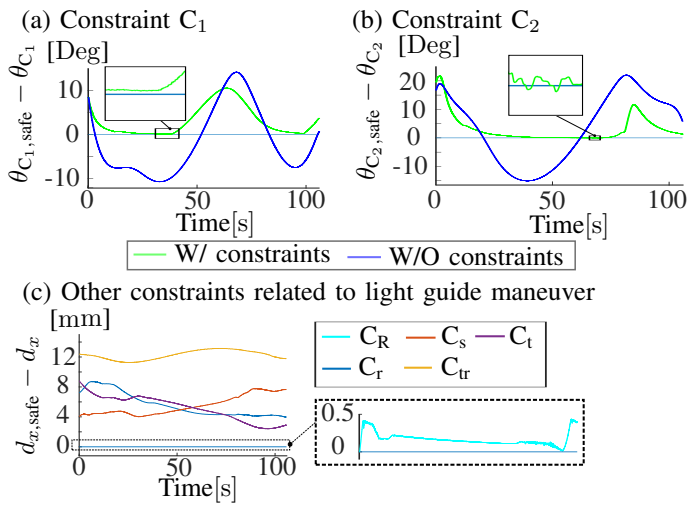

\caption{\label{fig:conical_constraint_result}The values related to the proposed
constraints for the light guide. A negative value means the constraint
is violated. (a) shows that the light guide satisfied the conical
constraint $\protect\cscope$ when it was autonomously controlled,
and $\protect\thetascope$ and $\protect\thetascopesafe$ are introduced
in \eqref{eq:C1_constraint}. (b) shows that the conical constraint
$\protect\cillu$ had a negligible violation of at most 0.00014~$\mathrm{rad}$
when it was autonomously controlled, and $\protect\thetaillu$ and
$\protect\thetaillusafe$ are introduced in \eqref{eq:C2_constraint}.
(c) shows that the light guide satisfied all other constraints when
it was autonomously controlled, and the distances are introduced in
Fig. \ref{fig:Vitreoretinal_constraints}.}
\end{figure}

Figs.~\ref{fig:conical_constraint_result}-(a), (b) show the distances
related to the conical constraints. A negative value means the constraint
is violated. The conical constraint $\cscope$ was always non-negative,
and $\cillu$ had a negligible violation of at most $0.0083\,\mathrm{deg}\ (0.00014\,\mathrm{rad})$.

Fig.~\ref{fig:conical_constraint_result}-(c) shows the distances
related to the constraints for the light guide maneuver (see Fig.~\ref{fig:Vitreoretinal_constraints}).
A negative value also means the constraint was violated. Fig.~\ref{fig:conical_constraint_result}-(c)
confirms that the light guide was autonomously controlled, satisfying
all the constraints.

\subsection{Simulation S1: Evaluation of robustness\label{subsec:Simulation}}

We conducted a simulation study to assess the robustness of our shadow-based
autonomous positioning strategy in the entire workspace. We tessellated
the workspace into a grid of 700 points as described in Fig.~\ref{fig:converge_iterations}
and evaluated whether the three-step algorithm converged properly
on each step and was successful.

\subsubsection*{Results and discussion}

\begin{figure}[tbh]
\centering
\def\svgwidth{230pt}
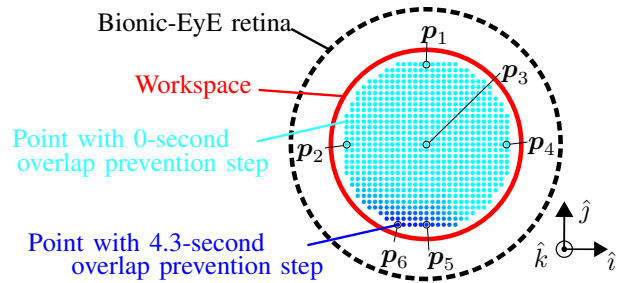

\caption{\label{fig:converge_iterations}The simulation result of our shadow-based
autonomous positioning for 700 points inside the workspace. The coordinates
are the same as in Fig.~\ref{fig:Workspace_and_RCM}. The controller
succeeded in positioning in all points. The color of each point represents
the time needed to complete the overlap prevention step\textcolor{blue}{.}
Six points $\protect\myvec p\in(\protect\myvec p_{1},\,\protect\myvec p_{2},\,\protect\myvec p_{3},\,\protect\myvec p_{4},\,\protect\myvec p_{5},\,\protect\myvec p_{6})$
are the representative points used in the following experiments. $\protect\myvec p_{6}$
is the point with the longest time for the overlap prevention step.}
\end{figure}

Our shadow-based autonomous positioning strategy properly converged
and succeeded in positioning the surgical instrument's tip in all
700 points.

In Fig.~\ref{fig:converge_iterations}, the color of each point represents
the time needed for the overlap prevention step to converge. In fact,
only some points around the lower-left part of the workspace needed
the overlap prevention step. Because of the geometry of the eye and
the trocar-point constraint, the instrument and its shadow tend to
overlap near the insertion point of the light guide (see Fig.~\ref{fig:Workspace_and_RCM}).
The longest time needed to complete the overlap prevention step was
approximately $4.3\,\mathrm{s}$ at most.

\subsection{Experiment E2: Evaluation using only the robot's kinematic models}

To evaluate our three-step shadow-based autonomous positioning strategy
with the physical robotic system, we commanded the robot to move to
representative points in the retina. We chose the six points $\myvec p\in(\myvec p_{1},\,\myvec p_{2},\,\myvec p_{3},\,\myvec p_{4},\,\myvec p_{5},\,\myvec p_{6})$
described in Fig.~\ref{fig:converge_iterations}, including the point
that needed the most iterations to complete the overlap prevention
step in the simulation study, $\myvec p_{6}$.

\subsubsection*{Results and discussion}

\begin{figure}[tbh]
\centering
\def\svgwidth{250pt}
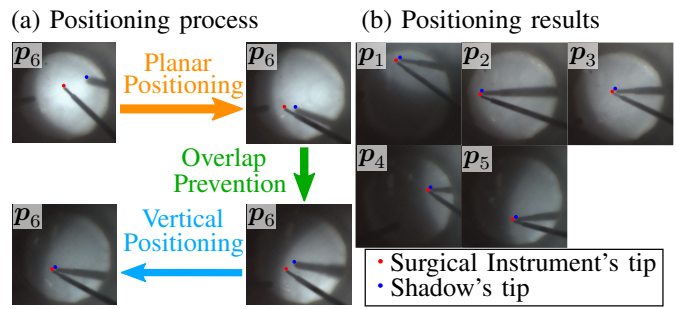

\caption{\label{fig:autonomous_experimental}The experimental results of the
evaluation of the entire process of our shadow-based positioning method.
(a) shows how the relative position of the surgical instrument and
its shadow changed in the microscopic view. (b) shows the appearances
after the proposed positioning to the representative points on the
Bionic-EyE retina.}
\end{figure}

Fig.~\ref{fig:autonomous_experimental}-(a) shows how the shadow
behaved in the microscopic view through the entire shadow-based autonomous
positioning process for $\myvec p_{6}$. We can see that the overlap
prevention step made the shadow easier to distinguish by separating
it from the shaft of the surgical instrument. In addition, the controller
succeeded in maintaining the tip of the shadow inside the microscopic
view during the whole process.

Fig.~\ref{fig:autonomous_experimental}-(b) shows the microscopic
view after the positioning process for the representative points,
$\myvec p$. At all points, it is shown that our proposed strategy
and controller can guide the shadow to a position where the shadow's
tip can be used as a cue to position the surgical instrument's tip
on the retina.

\subsection{Experiment E3: Integration with image processing}

In the following experiment, we integrated our system with an image-processing
strategy and re-evaluated our three-step shadow-based autonomous positioning.
For this purpose, we modified the Bionic-Eye's retina to be able to
detect the contact between the surgical instrument's tip and the retina,
trained a data-driven single-shot UNet-based \cite{ronnebergerUNetConvolutionalNetworks2015}
instrument tracking strategy, and slightly modified the three-step
positioning switching conditions based on that extra information.
These three modifications are described in detail as follows.

\subsubsection{Detection of contact}

\begin{figure*}[tbh]
\centering
\def\svgwidth{500pt} 
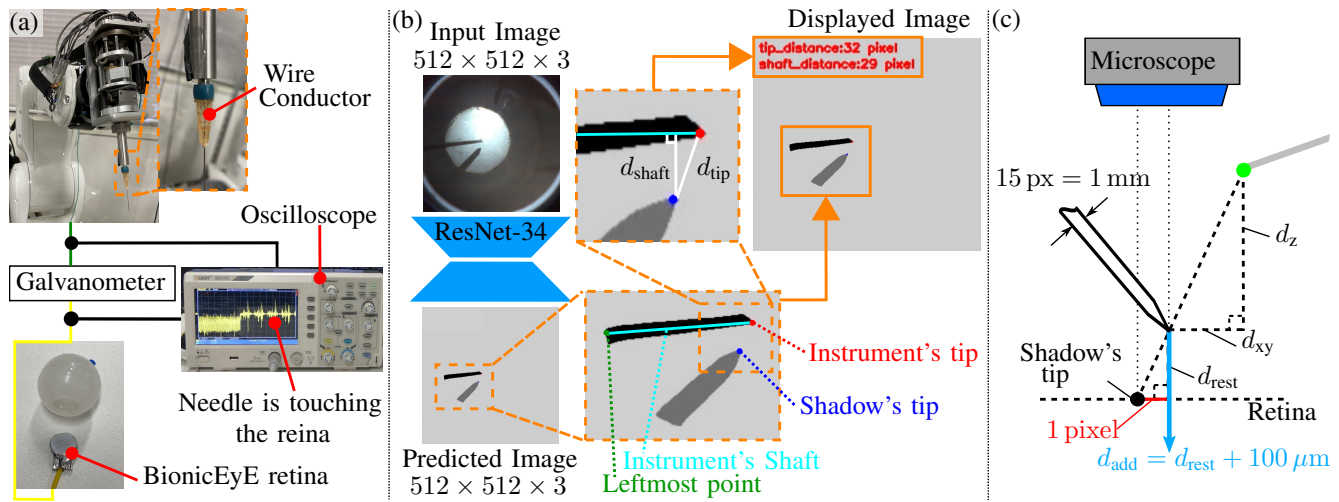

\caption{\label{fig:image_processing_integration} Integration with image processing.
(a) shows how we detect the contact between the surgical instrument's
tip and the retina. The detection can be observed as the change of
the waveform. (b) shows the flow of calculating the distances $\protect\shaftdis$
and $\protect\tipdis$ from the microscopic images. The shapes of
the surgical instrument and the shadow are predicted using U-Net.
Then, the distances are calculated based on the predicted image. (c)
shows how much controller pushes the surgical instrument, $\protect\additionaldis$,
after vertical positioning step and the geometrical relation ship
to calculate the distance.}
\end{figure*}

To detect the contact between the surgical instrument's tip and the
retina, we modified the setup as shown in Fig.~\ref{fig:image_processing_integration}-(a).
We used a stainless steel needle and covered the retina with an aluminum
sheet. We connected those to an oscilloscope. We can confirm the contact
as a change in the waveform on the oscilloscope.

We used a stainless steel needle in this experiment because the forceps
used in the previous experiments did not conduct current. Moreover,
we painted the retina white to reflect the color of the retina used
in the other experiments.

\subsubsection{Image processing}

To obtain the distances $\shaftdis$ and $\tipdis$ in the microscopic
view, we implemented a UNet-based image-processing strategy. As shown
in Fig~\ref{fig:image_processing_integration}-(b), the microscopic
images ($2048\times2048\times3$) were obtained using a camera (STC-MCCM401U3V,
Omron-Sentech, Japan). The images were resized to $512\times512\times3$
and input to a segmentation model based on UNet \cite{ronnebergerUNetConvolutionalNetworks2015}.
The output of the network is the three-class pixel-wise semantic segmentation
of the input image, classified into background, surgical instrument,
or shadow. Using the semantic segmented image, we use the edge points
of the predicted shapes as the tips, and the line that connects the
surgical instrument's tip and the leftmost point of the surgical instrument
is defined as the shaft. Finally, using these positions, $\shaftdis$
and $\tipdis$ are calculated.

\begin{figure*}[tbh]
\centering
\def\svgwidth{500pt} 
\import{fig/}{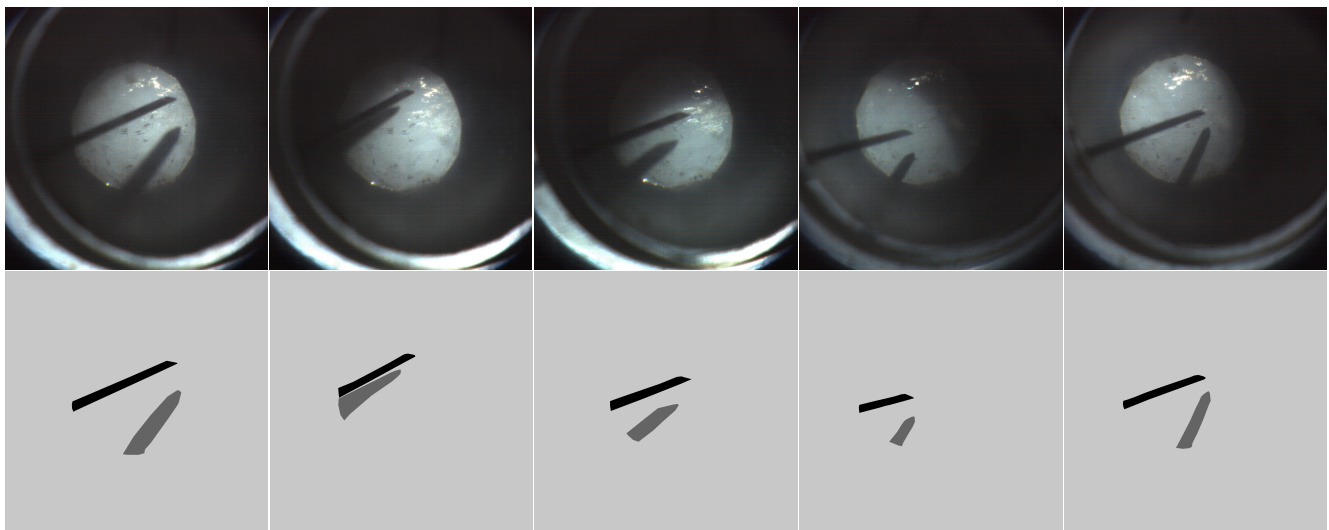}

\caption{\label{fig:ground-truth} Samples of manually segmented images.}
\end{figure*}

\begin{figure}[tbh]
\centering
\def\svgwidth{250pt}
\begingroup%
  \makeatletter%
  \providecommand\color[2][]{%
    \errmessage{(Inkscape) Color is used for the text in Inkscape, but the package 'color.sty' is not loaded}%
    \renewcommand\color[2][]{}%
  }%
  \providecommand\transparent[1]{%
    \errmessage{(Inkscape) Transparency is used (non-zero) for the text in Inkscape, but the package 'transparent.sty' is not loaded}%
    \renewcommand\transparent[1]{}%
  }%
  \providecommand\rotatebox[2]{#2}%
  \newcommand*\fsize{\dimexpr\f@size pt\relax}%
  \newcommand*\lineheight[1]{\fontsize{\fsize}{#1\fsize}\selectfont}%
  \ifx\svgwidth\undefined%
    \setlength{\unitlength}{994.87498847bp}%
    \ifx\svgscale\undefined%
      \relax%
    \else%
      \setlength{\unitlength}{\unitlength * \real{\svgscale}}%
    \fi%
  \else%
    \setlength{\unitlength}{\svgwidth}%
  \fi%
  \global\let\svgwidth\undefined%
  \global\let\svgscale\undefined%
  \makeatother%
  \begin{picture}(1,1.04642906)%
    \lineheight{1}%
    \setlength\tabcolsep{0pt}%
    \put(0,0){\includegraphics[width=\unitlength,page=1]{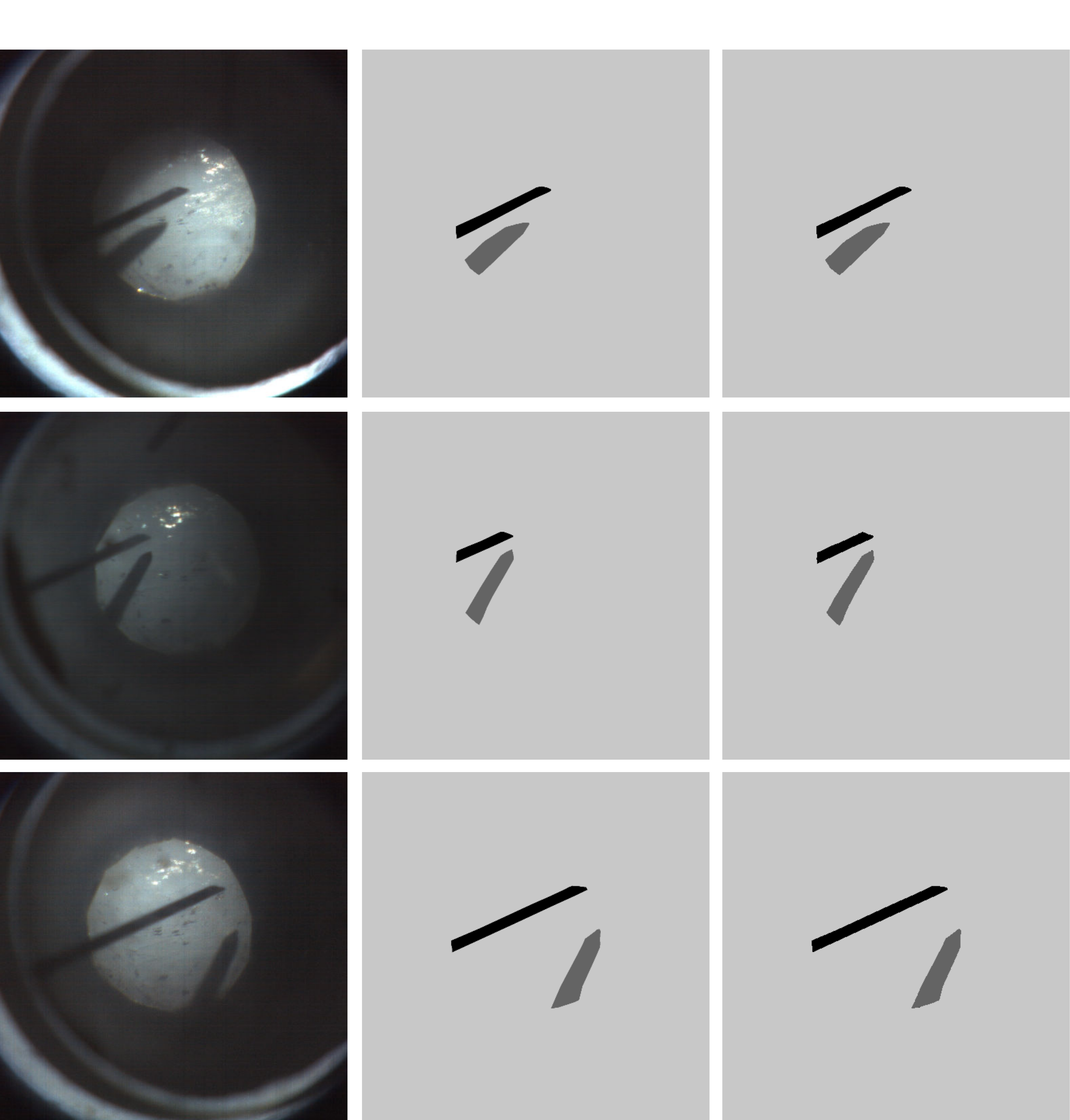}}%
    \put(0.00329238,1.02435797){\color[rgb]{0,0,0}\makebox(0,0)[lt]{\lineheight{1.25}\smash{\begin{tabular}[t]{l}Microscopic Image\end{tabular}}}}%
    \put(0.37718299,1.02225078){\color[rgb]{0,0,0}\makebox(0,0)[lt]{\lineheight{1.25}\smash{\begin{tabular}[t]{l}Ground Truth\end{tabular}}}}%
    \put(0.70726352,1.02102615){\color[rgb]{0,0,0}\makebox(0,0)[lt]{\lineheight{1.25}\smash{\begin{tabular}[t]{l}Predicted Image\end{tabular}}}}%
  \end{picture}%
\endgroup%

\caption{\label{fig:test_image} Results of the semantic segmentation of three
real microscopic images using the trained network.}
\end{figure}

The segmentation model was implemented using a PyTorch library \cite{Yakubovskiy:2019},
using a ResNet-34 pre-trained on ImageNet as encoder. To specialize
the network to our application, we collected 150 microscopic images
and manually segmented them to create ground truth images as shown
in Fig.~\ref{fig:ground-truth}. Furthermore, we used common augmentation
strategies \cite{info11020125}: blur, additive noise, image compression,
rotation, optical distortion, and random brightness, and contrast.
Considering all augmentations, we had 9000 images for training, 500
images for validation, and 500 images for testing.

The network was trained using dice loss, Adam optimizer with a learning
rate of $0.0001$, and batch size of $4$. We trained the network
on a single NVIDIA GeForce RTX 3070 GPU. The dice loss and Intersection
over Union (IoU) values on the test set were $0.0063$ and $0.9875$,
respectively. Fig.~\ref{fig:test_image} shows representative examples
of the semantic segmentation.

\subsubsection{Integration of the three-step algorithm with image processing\label{subsec:Integration-of-the}}

When $\tipdis$ equals $1\,\mathrm{px}$ in the microscopic view during
the vertical positioning step (see Section~\ref{subsec:Distance-information}),
it means that the instrument's tip is almost at the retina, as shown
in Fig~\ref{fig:image_processing_integration}-(c). After reaching
that point, we rely on the kinematic model to slowly move the surgical
instrument's tip downward.

The distance between the surgical instrument's tip and the retina
after the vertical positioning step, $\restdis$, can be calculated
using the geometrical relationship described in Fig.~\ref{fig:image_processing_integration}-(c),
as follows
\begin{align*}
d_{\text{rest}}\,[\mathrm{\mu m}]= & 1\,[\mathrm{px}]\times\frac{d_{z}\,[\mathrm{\mu m}]}{d_{xy}\,[\mathrm{\mu m}]}\times\frac{1}{\mathrm{convert}\,[\mathrm{px/\mu m}]}\text{,}
\end{align*}
where $d_{xy}$ and $d_{z}$ are, respectively, the $xy$ and $z$
components of the relative position of the light guide's tip and the
instrument's tip. In addition, $\mathrm{convert}=0.015\,[\mathrm{px/\mu m}]$
is calculated using as reference the diameter of the $1$-$\mathrm{mm}$-needle
in the microscopic image. Lastly, to ensure positioning, we move the
instrument $\additionaldis=\restdis+100\,\mathrm{\mu m}$ further.
We call this as additional positioning, for convenience.

\subsubsection{Validation}

As a proof-of-concept integration validation, we integrated our proposed
control strategy with image processing and evaluated our shadow-based
autonomous positioning. For this, we conducted autonomous positioning
for the six representative points used in \textbf{S1 }and\textbf{
E2,} $\myvec p\in(\myvec p_{1},\,\myvec p_{2},\,\myvec p_{3},\,\myvec p_{4},\,\myvec p_{5},\,\myvec p_{6})$
on the retina. The positioning for each point was repeated five times.

\subsubsection*{Results and Discussion}

\begin{table}[tbh]
\caption{\label{tab:values_of_additional_distance}Values of $\protect\additionaldis\,[\mathrm{\mu m}]$
for each point in each trial}

\begin{centering}
\begin{tabular}{|c|c|c|c|c|c|c|c|}
\hline 
 & T1 & T2 & T3 & T4 & T5 & Mean & Std. Dev.\tabularnewline
\hline 
\hline 
$\quat p_{1}$ & $246^{\dagger}$ & $244^{\dagger}$ & $241^{\diamondsuit}$ & $242^{\dagger}$ & $240^{\dagger}$ & $243$ & $2$\tabularnewline
\hline 
$\quat p_{2}$ & $281^{\dagger}$ & $286^{\diamondsuit}$ & $289^{\dagger}$ & $289^{\dagger}$ & $290^{\dagger}$ & $287$ & $4$\tabularnewline
\hline 
$\quat p_{3}$ & $229^{\dagger}$ & $230^{\dagger}$ & $230^{\dagger}$ & $230^{\dagger}$ & $239^{\diamondsuit}$ & $232$ & $4$\tabularnewline
\hline 
$\quat p_{4}$ & $229^{\dagger}$ & - & $238^{\diamondsuit}$ & - & $236^{\diamondsuit}$ & $234$ & $5$\tabularnewline
\hline 
$\quat p_{5}$ & $349^{\dagger}$ & $351^{\star}$ & $354^{\dagger}$ & $354^{\dagger}$ & $352^{\dagger}$ & $352$ & $3$\tabularnewline
\hline 
$\quat p_{6}$ & $392^{\dagger}$ & $366^{\star}$ & $406^{\dagger}$ & - & $408^{\dagger}$ & $393$ & $19$\tabularnewline
\hline 
\end{tabular}
\par\end{centering}
\medskip{}

{\small{}Contact with the retina was verified using the oscilloscope.}{\small\par}

{\small{}$\diamond$ No contact}{\small\par}

{\small{}$\star$ Contact during the vertical positioning step}{\small\par}

{\small{}$\dagger$ Contact during the additional positioning step}{\small\par}

{\small{}- Force-quit (technical problems related to the proof-of-concept
integration, stopped manually for safety) }{\small\par}
\end{table}

Table~\ref{tab:values_of_additional_distance} shows the values of
$\additionaldis\,[\mathrm{\mu m}]$ and how the autonomous positioning
finished for each point in each trial. In the trials with $\dagger$,
the surgical instrument's tip properly touched the retina during the
additional positioning step.

\begin{figure}[tbh]
\centering
\def\svgwidth{250pt} 
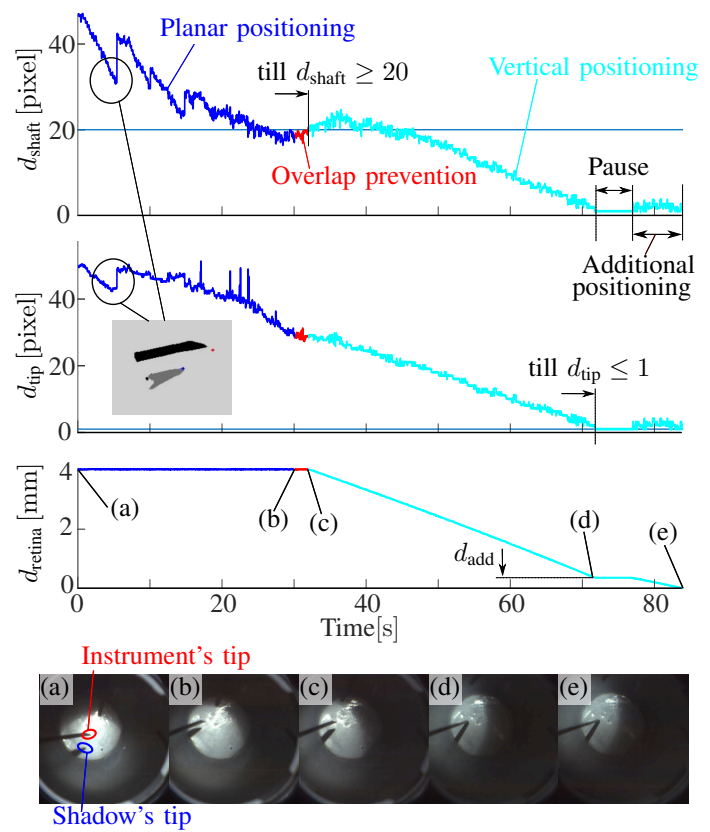

\caption{\label{fig:result_distance}The system behavior during shadow-based
autonomous positioning for $\protect\quat p_{5}$. The distance $\protect\retinadis$
is the distance between the surgical instrument's tip and the retina,
calculated using the kinematic models of the robots, with zero at
the end of positioning. The pictures below show the microscopic images
at each stage.}
\end{figure}

In the cases marked with $\diamond$, the tip did not touch the retina
even after the additional positioning step. On the other hand, in
cases marked with $\star$, the surgical instrument's tip touched
the retina before the additional positioning step. This can be caused
by several factors, given that the scale of all objects in this experiment
is very small. For instance, the surface of the retina that we prepared
for this experiment was made to be convenient for detecting contact
but does not have uniform height and placement. Moreover, the current
resolution of the microscopic image is approximately $1\,\mathrm{px}=70\,\mathrm{\mu m}$
and must be improved in future work. This also affects the image-processing
strategy to find the instrument's tip (e.g. the noise in Fig.~\ref{fig:result_distance}),
which should be further improved.

As for the cases of $\quat p_{4}$ with -, the needle used for this
experiment was too short. A longer needle will be used in further
validations.

Fig.~\ref{fig:result_distance} shows the system behavior during
shadow-based autonomous positioning for $\quat p_{5}$. We can confirm
that the controller switched between the steps according to the thresholds
$\overlapthreshold=20\,\mathrm{px}$ and $\verticalthreshold=1\,\mathrm{px}$.
The total computational time from acquiring the image to finishing
calculating the distances was $0.093\,\mathrm{s}$. The distance $\retinadis$
is calculated using the robot's kinematic parameters after the experiment.

\section{Conclusion}

In this work, a novel three-step shadow-based autonomous positioning
strategy for vitreoretinal tasks was proposed. This method requires
bimanual control that guarantees the visibility of the shadow of the
surgical instrument's tip in the microscopic view. To achieve this,
we derived new conical VFIs, used to autonomously move the light guide
with respect to the surgical instrument. The results of the experimental
and simulation studies confirm the feasibility of our positioning
strategy in the entire workspace. In future work, we aim to improve
image processing and use visual feedback for more accurate positioning.
Furthermore, the proposed autonomous bimanual control of the light
guide has the potential to improve the efficiency of surgical procedures
and lead to new surgical techniques by setting the one hand of the
surgeon free. We also aim to research how the proposed control can
be used in collaborative work with the surgeons in future work.

\bibliographystyle{IEEEtran}
\bibliography{TMRB}

\end{document}